\documentclass[10pt,twocolumn,letterpaper]{article}

\usepackage[algorithms]{wacv}
\usepackage{times}
\usepackage{epsfig}
\usepackage{graphicx}
\usepackage{booktabs}

\usepackage{tikz}
\usepackage{comment}
\usepackage{color}
\usepackage{graphicx}
\usepackage{amsmath}
\usepackage{amssymb}
\usepackage{booktabs}
\usepackage{multirow, boldline}
\usepackage{colortbl}
\usepackage{stmaryrd}
\usepackage{scalerel}
\usepackage{enumitem}
\usepackage{pifont}
\usepackage[export]{adjustbox}
\usepackage{makecell}
\usepackage{mathabx}
\usepackage{xfrac}
\usepackage{bm}

\usepackage[pagebackref,breaklinks,colorlinks]{hyperref}

\usepackage[capitalize]{cleveref}
\usepackage{ulem}

\begin{document}

\Crefname{section}{Section}{Sections}
\Crefname{table}{Table}{Tables}

\definecolor{cite_color}{RGB}{7, 150, 255}
\definecolor{text_blue}{RGB}{108, 142, 191}
\definecolor{text_orange}{RGB}{215, 155, 0}
\definecolor{text_green}{RGB}{130, 179, 102}
\definecolor{rred}{RGB}{216, 27, 96}
\definecolor{ogreen}{RGB}{106, 122, 191}

\newcommand{\STAB}[1]{\begin{tabular}{@{}c@{}}#1\end{tabular}}

\newcommand{\dir}[2]{#1 \shortrightarrow #2}

\newcommand{\THVc}[1]{\textcolor{olive}{[\textbf{Thv}: #1]}}
\newcommand{\THV}[1]{\textcolor{olive}{#1}}
\newcommand{\RCc}[1]{\textcolor{purple}{[\textbf{RC}: #1]}}
\newcommand{\RC}[1]{\textcolor{purple}{#1}}
\newcommand{\ILc}[1]{\textcolor{blue}{[\textbf{IL}: #1]}}
\newcommand{\IL}[1]{\textcolor{blue}{#1}}

\newcommand{\todo}[1]{\textcolor{red}{#1}}
\newcommand{\paragr}[1]{\noindent\textbf{#1}~} 
\newcommand{\resstd}[2]{#1 $\pm$ #2}

\newcommand{\corr}[1]{#1}
\newcommand{\revcorr}[2]{\textcolor{red}{\sout{#1}}\textcolor{blue}{#2}}

\newcommand{\rtb}[1]{\rotatebox{90}{#1}}   %
\newcolumntype{C}[1]{>{\centering\arraybackslash}p{#1}}	
\newcolumntype{L}[1]{>{\raggedright\arraybackslash}p{#1}}
\newcolumntype{R}[1]{>{\raggedleft\arraybackslash}p{#1}}

\newcommand{\SD}{`${S\text{-}D}$'}
\newcommand{\SDN}{`${S\text{-}D\text{-}N}$'}
\newcommand{\SDNE}{`${S\text{-}D\text{-}N\text{-}E}$'}

\newcommand{\mainblock}[0]{multi-Task Exchange Block}
\newcommand{\mainblockshort}[0]{mTEB}

\newcommand{\qualdata}[0]{{Cityscapes}}

\newcommand{\oc}[1]{\cellcolor{gray!10}}    %
\newcommand{\bc}[1]{\cellcolor{yellow!10}}  %

\newcolumntype{?}{!{\vrule width 1pt}}

\definecolor{xblue}{RGB}{189,217,238}
\definecolor{xorange}{RGB}{248,203,173}
\definecolor{xgreen}{RGB}{197,224,180}
\definecolor{xgray}{RGB}{217,217,217}
\definecolor{xpurple}{RGB}{215,170,215}

\newcommand{\hsem}{\cellcolor{xblue}}
\newcommand{\hdepth}{\cellcolor{xorange}}
\newcommand{\hnormal}{\cellcolor{xgreen}}
\newcommand{\hdelta}{\cellcolor{xgray}}
\newcommand{\hedge}{\cellcolor{xpurple}}

\newcommand{\param}[1]{#1}
\newcommand{\perf}[2]{#1\scaleto{\;\pm #2}{3pt}}
\newcommand{\uperf}[2]{\underline{#1}\scaleto{\;\pm #2}{3pt}}
\newcommand{\bperf}[2]{\textbf{#1}\scaleto{\;\pm #2}{3pt}}
\newcommand{\perfn}[1]{#1}

\newcommand{\rel}[1]{\scaleto{#1}{4pt}}
\newcommand{\urel}[1]{\underline{\rel{#1}}}
\newcommand{\brel}[1]{\textbf{\rel{#1}}}

\newcommand{\darr}[0]{$\downarrow$}
\newcommand{\uarr}[0]{$\uparrow$}

\newcommand{\dmtl}[1]{\Delta\textsubscript{#1}}

\newcommand{\tick}[0]{\ding{51}}
\newcommand{\cross}[0]{\ding{55}}

\newcommand{\threewaysPADNet}{$\text{3-ways}_\text{\scalebox{0.7}{{PAD-Net}}}$}
\newcommand{\threewaysOurs}{$\text{3-ways}_\text{\scalebox{0.7}{{mTEB}}}$}

\newcommand{\vk}[0]{VKITTI2}
\newcommand{\sy}[0]{Synthia}
\newcommand{\cs}[0]{Cityscapes}
\newcommand{\nyu}[0]{NYUDv2}

\definecolor{csroad}{RGB}{128, 64, 128}
\definecolor{csside}{RGB}{244, 35, 232}
\definecolor{csbuild}{RGB}{70, 70, 70}
\definecolor{cswall}{RGB}{102, 102, 156}
\definecolor{csfence}{RGB}{190, 153, 153}
\definecolor{cspole}{RGB}{153, 153, 153}
\definecolor{cslight}{RGB}{250, 170, 30}
\definecolor{cssign}{RGB}{220, 220, 0}
\definecolor{csveg}{RGB}{107, 142, 35}
\definecolor{cssky}{RGB}{70, 130, 180}
\definecolor{csperson}{RGB}{220, 20, 60}
\definecolor{csrider}{RGB}{255, 0, 0}
\definecolor{cscar}{RGB}{0, 0, 142}
\definecolor{csbus}{RGB}{0, 60, 100}
\definecolor{csmbike}{RGB}{0, 0, 230}
\definecolor{csbike}{RGB}{119, 11, 32}

\title{Cross-task Attention Mechanism for Dense Multi-task Learning}

\author{Ivan Lopes\textsuperscript{1}
\and
Tuan-Hung Vu\textsuperscript{1,2}
\and
Raoul de Charette\textsuperscript{1}
\and
\textsuperscript{1}Inria \quad \textsuperscript{2}Valeo.ai\\ \\
\url{https://github.com/astra-vision/densemtl}
}

\maketitle

\begin{abstract}
Multi-task learning has recently become a promising solution for comprehensive understanding of complex scenes.
With an appropriate design, multi-task models can not only be memory-efficient but also favour the exchange of complementary signals across tasks. In this work, we jointly address 2D semantic segmentation, and two geometry-related tasks, namely dense depth, surface normal estimation as well as edge estimation showing their benefit on several datasets.
We propose a novel multi-task learning architecture that exploits pair-wise cross-task exchange through \textit{correlation-guided attention} and \textit{self-attention} to enhance the average representation learning for all tasks.
We conduct extensive experiments on three multi-task setups, showing the benefit of our proposal in comparison to competitive baselines in both synthetic and real benchmarks.
We also extend our method to the novel multi-task unsupervised domain adaptation setting. Our code is open-source.

\end{abstract}

\section{Introduction}
\label{sec:intro}
Recent advances in deep neural network architectures~\cite{he2016deep,ioffe2015batch} and efficient optimization techniques~\cite{bottou2010large,kingma2014adam} constantly set higher accuracy in scene understanding tasks, demonstrating great potentials for autonomous applications.
Nevertheless, the majority of the literature focuses on pushing performance of single-tasks, being either semantic tasks like segmentation~\cite{chen2017deeplab,hoyer2021ways} and detection~\cite{wu2019detectron2}, or geometrical tasks like depth/normal estimation~\cite{li2015depth,godard2019digging}.
Very few have paid attention to a more comprehensive objective of joint semantics and geometry understanding, while in practice, this is desirable in critical applications such as robotics and autonomous driving.
In those systems, we would expect to have cooperative synergies among all tasks, \ie tasks should be processed together in one unified system rather than separately.
Arguably, promoting such synergies could bring mutual benefit to all tasks involved.
For example, disruptive changes in depth maps may signal semantic boundaries in segmentation maps; while pixels of some semantic classes, like ``road'', may share similar surface normals.

To this end, multi-task learning (MTL)~\cite{caruana1997multitask,kokkinos2017ubernet,tosi2020distilled} has become a promising solution as it seeks to learn a unified model that excels in average on \textit{all} tasks. %
A common MTL design is to have a large amount of parameters shared among tasks while keeping certain private parameters for individual ones; information is exchanged via shared parameters, making possible the synergy flow.
Some recent works~\cite{bruggemann2021exploring,vandenhende2020mtinet,xu2018padnet} focus on new multi-modal modules that boost tasks interactions.
One important advantage of such MTL models is memory-efficiency thanks to the shared parts.
In this work, we adopt the same principle and design our model with shared encoder and dedicated task decoders.

\begin{figure}[!t]
	\centering
	\includegraphics[width=0.98\linewidth,height=3.0cm]{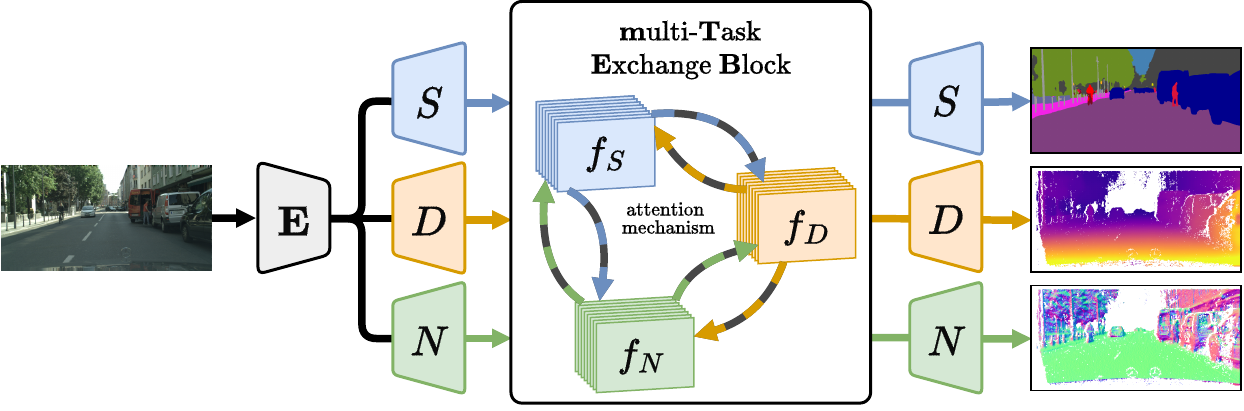}
	\caption{\textbf{Overview of our MTL framework.} The three tasks semantic Segmentation, Depth regression and Normal estimation share an encoder $\textbf{E}$. The task-specific decoders $S$, $D$ and $N$ exchange information in the ``multi-Task Exchange Block'' (mTEB) via an attention-based mechanism, resulting in refined features to produce final predictions.
	}
	\label{fig:teaser}
\end{figure}

Recent MTL approaches come with different solutions to improve the information exchanged or distilled across tasks, often known as \textit{multi-modal distillation}.
PAD-Net~\cite{xu2018padnet} and MTI-Net~\cite{vandenhende2020mtinet} have proven the effectiveness of self-attention~\cite{vaswani2017attention} for MTL, \ie a mechanism to self-discover most attentive signals of each task-specific feature for the other tasks.
Differently, ATRC~\cite{bruggemann2021exploring} advocates to better look at pair-wise task similarity to guide multi-modal distillation. We partially follow the same direction as in ATRC with our \textit{correlation-guided attention}, in which we use correlations between task-specific features to guide the construction of exchanged messages.
Further, we propose an unified mechanism that combines two different attentions, \textit{correlation-guided attention} and \textit{self-attention}, via a learnable channel-wise weighting scheme.

In summary, we propose a novel multi-modal distillation design for MTL, relying on pair-wise cross-task attentions mechanisms (coined xTAM, \cref{sec:xtam}) combined into a multi-task exchange block (coined mTEB, \cref{sec:meth-mtl-refblock}).
We address three critical tasks for outdoor scene understanding: semantic segmentation, dense depth estimation and surface normal estimation.
In extensive experiments on multiple benchmarks, our proposed framework outperforms competitive MTL baselines (\cref{sec:exp_mtl}).
With our new multi-task module, we also report improvement in semantic segmentation on Cityscapes where self-supervised depth plays as the auxiliary task~\cite{hoyer2021ways}.
Empirically, we showcase the merit of our proposed multi-task exchange in a new setting of MTL unsupervised domain adaptation (MTL-UDA, \cref{sec:exp_mtlda}), outperforming all baselines.

\newcommand{\relparagr}[1]{\noindent\textit{\underline{#1}}}

\section{Related Work}
\label{sec:related}

\relparagr{Multi-task learning.} In the early days of neural networks, Caruana~\cite{caruana1997multitask} introduces the idea of multi-task learning with \textit{hard parameter sharing}, \ie some parameters are shared between all tasks, some are dedicated to separate tasks.
In the same spirit, UberNet~\cite{kokkinos2017ubernet} features a deep architecture to jointly address a large number of low-, mid- and high-level tasks.
Zamir \textit{et al.}~\cite{zamir2018taskonomy} conduct a large-scale study with 26 tasks on four million images of indoor scenes, studying the dependencies and transferabilities across tasks.
While those seminal works show great potential of multi-task learning, they do note a few challenges, most notably the \textit{negative transfer} phenomenon that degrades performance of certain tasks when learned jointly~\cite{kokkinos2017ubernet,kendall2018multi}.
Some works~\cite{kendall2018multi,chen2018gradnorm} reason that negative transfer is due to the imbalance of multi-task losses and introduce mechanisms to subsequently weight these individual loss terms.
Kendall~\textit{et al.}~\cite{kendall2018multi} propose to weight multiple losses by estimating the homoscedastic uncertainty of each task.
Chen~\textit{et al.}~\cite{chen2018gradnorm} introduce GradNorm, an algorithm that helps dynamically tuning gradient magnitudes such that the learning pace of different tasks are balanced.
Differently, Sener and Koltun~\cite{sener2018multi} cast multi-task learning as multi-objective optimization where the goal is to find Pareto-optimal solutions.

\relparagr{Cross-task mechanisms.} Closer to our work are methods focusing on improving exchange or distillation across tasks~\cite{xu2018padnet,zhang2019patternaffinitive,vandenhende2020mtinet,bruggemann2021exploring}; the main idea is that each task could benefit from different yet complementary signals from the others.
Inspired from the success of visual attention in perception tasks~\cite{xu2015show,vaswani2017attention,fu2019dual}, PAD-Net~\cite{xu2018padnet} uses an attention mechanism to distill information across multi-modal features.
In MTI-Net~\cite{vandenhende2020mtinet}, Vandenhende~\textit{et al.} extend PAD-Net with a multi-scale solution to better distill multi-modal information.
Zhang~\textit{et al.}~\cite{zhang2019patternaffinitive} propose to aggregate affinity maps of all tasks, which is then ``diffused'' to refine task-specific features.
PSD \cite{zhou2020psd}~mines and propagates patch-wise affinities via graphlets, instead we model the interactions with attention mechanisms.
Bruggemann~\textit{et al.}~\cite{bruggemann2021exploring} introduce ATRC to enhance multi-modal distillation with four relational context types based on different types and levels of attention.
Similar to ATRC, our method also exploits pairwise task similarity to refine task-specific features.
Differently though, we advocate to combine the pairwise similarity with cross-task self-attention via a learnable weighting scheme to learn the refined residuals that complement the original task-specific features.

Recent efforts seek for label-efficient learning paradigms to train models for urban scene understanding.
Tosi~\textit{et al.}~\cite{tosi2020distilled} introduces real-time multi-task network based on knowledge distillation and self-supervision training.
Hoyer~\textit{et al.}~\cite{hoyer2021ways} propose a novel architecture and different strategies to improve semantic segmentation with self-supervised monocular depth estimation.
One finding in~\cite{hoyer2021ways} is that the MTL module of PAD-Net~\cite{xu2018padnet} complements other self-supervised strategies and further improve segmentation performance.
On this regard, we study the effect brought by our proposed module in this particular setup.

\section{Method}
\label{sec:method}
In multi-task learning, the aim is to optimize a set of $n$ tasks: $i \in \{{T}_1, ... , {T}_n\}$ while seeking a \textit{general} good performance -- as opposed to favoring a single task.
Our model takes an input image and makes $n$ predictions.
Generally, this is achieved with a shared encoder and separate task-specific decoders.
Here, we introduce a novel mechanism that enhances cross-task talks via features exchange, building on the intuition that each decoder discovers unique but complementary features due to its separate supervision~\cite{zamir2018taskonomy}.

We formulate in \cref{sec:xtam} a bidirectional cross-Task Attention Mechanism module, coined xTAM, taking as input a pair of task-specific features and returning two \textit{directional features}.
We then present in \cref{sec:mtl_overview} our complete MTL framework for scene understanding that encompasses our multi-Task Exchange Block (\mainblockshort{}, see \cref{fig:teaser}).

\subsection{xTAM: Bidirectional Cross-Task Attention Mechanism}
\label{sec:xtam}

\definecolor{figgreen}{RGB}{130, 179, 102}
\definecolor{figpurple}{RGB}{150, 115, 166}
\begin{figure*}[t]
	\centering
	\includegraphics[width=0.75\linewidth,trim=0.5cm 1cm 0.5cm 0.3cm, clip]{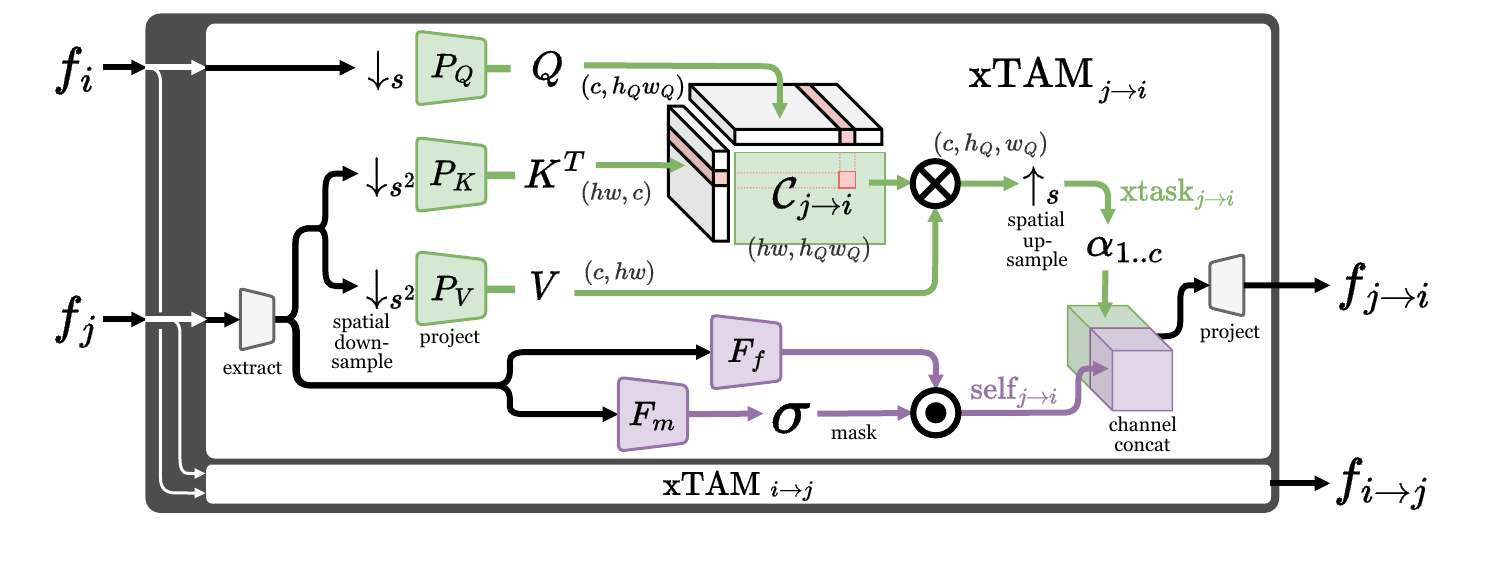}
	\caption{\textbf{Bidirectional Cross-task Attention (xTAM).}
		We enable information flow between task pairs ($i$, $j$) via the discovery of directional features, $f_{\dir{j}{i}}$ and $f_{\dir{i}{j}}$. Only $\text{xTAM}_{\dir{j}{i}}$ is detailed here. It relies on two attention mechanisms. First, a correlation-guided attention ({\color{figgreen}green}) to output $\textbf{xtask}_{\dir{j}{i}}$, the features from task~$j$ contributing to task~$i$. Second, a self-attention ({\color{figpurple}purple}) to discover complementary signal $\textbf{self}_{\dir{j}{i}}$ from $j$. 
Best viewed in color.
	}
	\label{fig:xtam}
\end{figure*}

Building upon recent works demonstrating task interactions \cite{standley2020tasks}, our module seeks to capture the shared pair-wise task knowledge while preserving their exclusive (non-shared) knowledge.
Our intuition is that we can exploit features from pairs of tasks denoted $(i, j)$ and self-discover the interactions from their correlation matrices for either directional interaction: $\dir{i}{j}$ and $\dir{j}{i}$. 
\cref{fig:xtam} illustrates our xTAM component which distils the knowledge between tasks, taking as input the task features $(f_i, f_j)$ and returning the \textit{directional features} for the other task, \ie $(f_{\dir{j}{i}}, f_{\dir{i}{j}})$.

For ease of speech we describe the directional $\text{xTAM}_{\dir{j}{i}}$ where task $j$ helps task $i$, but emphasize that xTAM being bidirectional, is made of both $\text{xTAM}_{\dir{j}{i}}$ and $\text{xTAM}_{\dir{i}{j}}$.

\subsubsection{Directional cross-task attention ($\text{xTAM}_{\dir{j}{i}}$).}
Considering here $i$ as our primary task, and $j$ as the secondary, we seek to estimate the features from task $j$ that can contribute to $i$.
We leverage two cross-task attentions: (i) a \textit{correlation-guided attention} which goal is to exploit cross-task spatial correlations to guide the extraction of contributing features of the secondary task to the primary task, and (ii) a \textit{self-attention} on the secondary task to self-discover complementary signals which are beneficial for the primary task.
Visualizations of the two attentions are colored in green and purple in \cref{fig:xtam}, respectively.
Note that each attention contributes differently to the primary task $i$ relying either on identification of shared $j$ and $i$ knowledge or on exclusive $j$ knowledge. We use 1x1 convolutional layers (gray blocks in \cref{fig:xtam}) for dimension compatibility.

\paragr{Correlation-guided attention.} To guide features, we rely on spatial correlations of task features (green blocks in \cref{fig:xtam}). In practice, we project \textit{downscaled} $f_i$ and $f_j$ features onto $d$-dimensional sub-spaces as in~\cite{vaswani2017attention} such that 
$K = P_{Q}({\downarrow_{s}}f_i)$, 
$Q = P_{K}({\downarrow_{s^2}}f_j)$, having ${\downarrow_{s}}$ the downscale operator with $s$ the scale factor, and $P_Q$, $P_K$ separate 1x1 convolutions.
The spatial-correlation matrix $\mathcal{C}_{\dir{j}{i}}$ is then obtained by applying a softmax on the matrix multiplication $K^{T}\times{}Q$ and normalizing with $\sqrt{d}$:

\begin{equation}
    \setlength{\abovedisplayskip}{3pt}
    \setlength{\belowdisplayskip}{3pt}
    \mathcal{C}_{\dir{j}{i}} = \text{softmax}\left(\frac{K^{T}\times Q}{\sqrt{d}}\right)\,.
    \label{eq:crosstaskji}
\end{equation}
where $d$ is the feature size \cite{vaswani2017attention}. Intuitively, $\mathcal{C}_{\dir{j}{i}}$ has high values where features from $i$ and $j$ are highly correlated and low values otherwise, which we use to weight features from~$f_j$. 
Subsequently, we obtain our correlation-guided attention features $\textbf{xtask}_{\boldsymbol{\dir{j}{i}}}$ by multiplying the correlation matrix $\mathcal{C}_{\dir{j}{i}}$ with the projected $j$ features:
\begin{equation}
    \setlength{\abovedisplayskip}{3pt}
    \setlength{\belowdisplayskip}{3pt}
	\textbf{xtask}_{\boldsymbol{\dir{j}{i}}} = \uparrow_{s}(V\times \mathcal{C}_{\dir{j}{i}})\,,
\end{equation}
with $V = P_{V}({\downarrow_{s^2}}f_j)$, $P_{V}$ the 1x1 projection, and ${\uparrow_{s}}$ the upsample operator.

\paragr{Self-attention.} %
We additionally employ a spatial attention \cite{xu2018padnet} which we denote as `self-attention' to contrast with the above which takes pairs of differing tasks.
Instead the following mechanism (purple blocks in \cref{fig:xtam}) takes features from $j$ alone and aims to extract private information from $f_j$ that are relevant for predicting task $i$:
\begin{equation}
\label{sa_eq}
    \setlength{\abovedisplayskip}{3pt}
    \setlength{\belowdisplayskip}{3pt}
    \textbf{self}_{\boldsymbol{\dir{j}{i}}} = F_f(f_j) \odot \sigma (F_m(f_j))\,,
\end{equation}
being our \textit{self-attention features}. Where $\odot$ is the element-wise multiplication, and $\sigma$ the sigmoid function.
Both $F_f$ and $F_m$ are convolutional layers which are supervised by the target task $i$ to learn to extract relevant information from features $f_j$.
The self-attention features $\textbf{self}_{\boldsymbol{\dir{j}{i}}}$ is defined as the point-wise multiplication between the features coming from $F_f$ and the dynamic mask provided by $F_m$.

\paragr{Directional feature.} To construct the final \textit{directional features} $f_{\dir{j}{i}}$ for the ${\dir{j}{i}}$ interaction, the two attention based feature maps are combined as:
\begin{equation}
    \setlength{\abovedisplayskip}{3pt}
    \setlength{\belowdisplayskip}{3pt}
    f_{\dir{j}{i}} = [\text{diag}(\alpha_1,...,\alpha_c) \times \textbf{xtask}_{\boldsymbol{\dir{j}{i}}} \,,\, \textbf{self}_{\boldsymbol{\dir{j}{i}}} ],
    \label{eqn:dir_feat}
\end{equation}
where $[.,.]$ is the channel-wise concatenation operation, and  $\alpha_{1..c}$ are learnable \textit{scalars} used to weight the $c$ channels of $\textbf{xtask}_{\boldsymbol{\dir{j}{i}}}$.
All $\alpha_{1..c}$ are initialized with $0$; learning will adaptively adjust the per-channel weighting.
Intuitively, cross-task exchange first starts with self-attention only, then gradually adjusts the $\alpha_{1..c}$ values to include some contribution from the correlation-guided attention.
This initialization strategy is important to stabilize training, especially here where we combine different types of attention.

Overall, the bidirectional xTAM block is made up of two directional blocks,  $\text{xTAM}_{\dir{j}{i}}$ and $\text{xTAM}_{\dir{i}{j}}$; it outputs both $f_{\dir{j}{i}}$ and $f_{\dir{i}{j}}$ for each $(i,\,j)$ task pair.

\subsubsection{Discussion.}
Our xTAM design is different from existing multi-modal distillation modules~\cite{xu2018padnet,vandenhende2020mtinet,bruggemann2021exploring}.
While PAD-Net~\cite{xu2018padnet} and MTI-Net~\cite{vandenhende2020mtinet} only consider cross-task self-attention to distill multi-modal information, ATRC~\cite{bruggemann2021exploring} models pair-wise task similarity, which shares a similar spirit to our correlation-guided attention.
Different to those, our xTAM learns to adaptively combine cross-task self-attention and correlation-guided attention using a learnable weighting scheme.
Of note, adopting the multi-scale strategy of MTI-Net~\cite{vandenhende2020mtinet} or having other pair-wise modules of ATRC~\cite{bruggemann2021exploring} is orthogonal to our xTAM.
There might be potential improvements by systematically combining those strategies with xTAM.
However, we focus here on giving an extensive study of xTAM design and applications; we leave such combinations for future investigation.

\definecolor{figred}{RGB}{234, 107, 102}
\subsection{Multi-task Learning Framework.} 
\label{sec:mtl_overview}

\begin{figure}[h!]
    \centering
    \includegraphics[width=\linewidth,trim=0 0.6cm 0 0, clip]{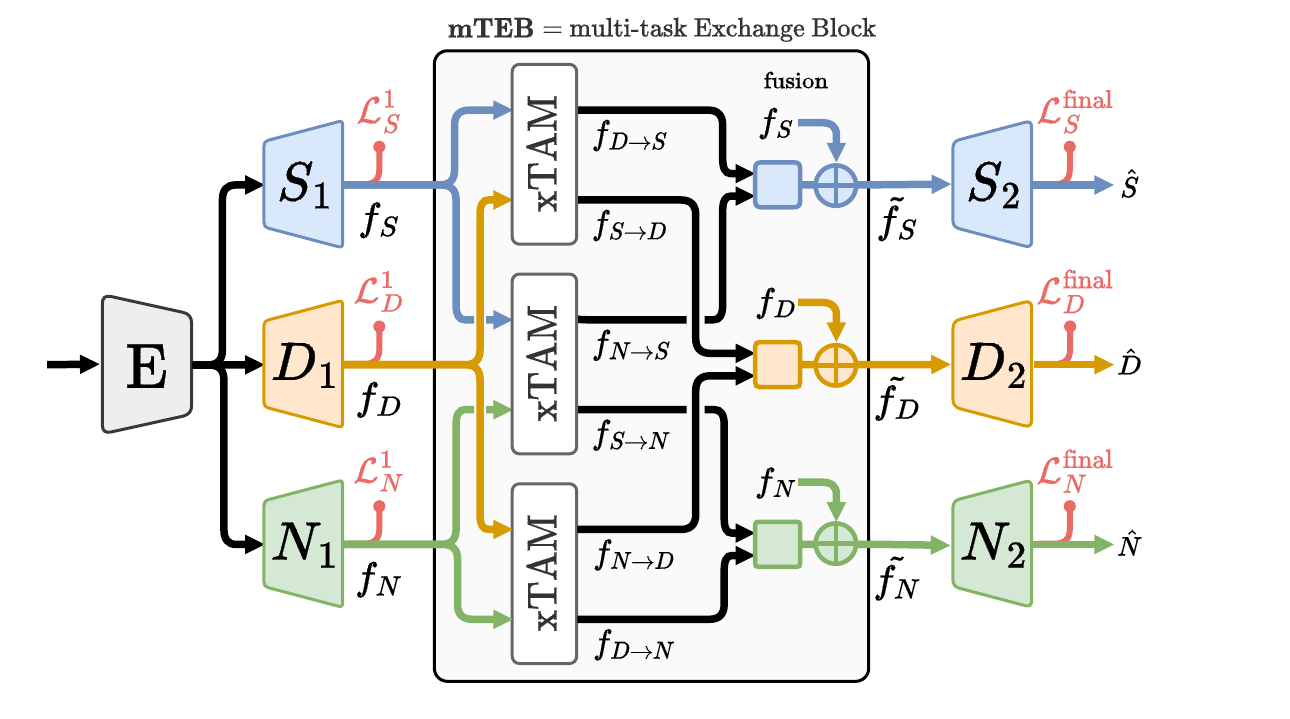}
    \caption{\textbf{Our cross-task MTL framework.} We jointly predict semantic segmentation ($S$,~{\color{text_blue}blue}), depth ($D$,~{\color{text_orange}orange}) and surface normal ($N$,~{\color{text_green}green}). All tasks share encoder ($\textbf{E}$) and have dedicated decoders (shown as trapezoids). Our cross-task attention module is inserted within a ``\mainblock'' (\mainblockshort{}) (\cref{sec:meth-mtl-refblock}) that takes task-specific features as inputs and returns refined features of the same kinds. Here, three xTAM modules are used to extract pairs of \textit{directional features}, where each represents the complementary signal from one task to another, \eg $f_{D\shortrightarrow S}$ encapsulates depth information ($D$) extracted to improve the segmentation task ($S$). Multi-modal \textit{directional features} are combined in the square blocks, resulting in multi-modal residuals for refinement. Task specific losses (shown in {\color{figred}red}) are applied before and after our \mainblockshort.}
    \label{fig:mtl}
\end{figure}

We focus on semantic and geometrical estimation tasks as they are critical for scene understanding especially outdoor environments.
\cref{fig:mtl} overviews our MTL framework having a unique encoder $\textbf{E}$ and illustrates the interactions taking place between the three different task decoders ($S$, $D$ and $N$) in a three-task setup: $T=\{\textbf{S}\text{egmentation},\allowbreak \textbf{D}\text{epth},\allowbreak \textbf{N}\text{ormal}\}$. We first describe our ``\mainblock'' (\mainblockshort{})  and then go over our training setup.

\subsubsection{\mainblock{} (\mainblockshort{}).}
\label{sec:meth-mtl-refblock} 
Our block %
can be inserted at any stage in the decoder, taking input task-specific features as inputs, here $(f_S,f_D,f_N)$, and outputting refined features, here  $(\tilde{f}_S,\tilde{f}_D,\tilde{f}_N)$.
The block consists of one bidirectional xTAM module (\cref{sec:xtam}) per task pair, each returning two directional features which are fused to the input task features.
Considering $i$ to be the primary task, we concatenate the $n-1$ \textit{directional features} $f_{\dir{j}{i}}|_{j\in T \setminus \{i\}}$, contributing towards task $i$, along the channel dimension.
The concatenated feature map is then processed with $F_{\dir{}{i}}\,$ (depicted as square blocks in \cref{fig:mtl}): a 1x1 convolution followed by a batch-normalization~\cite{ioffe2015batch} and ReLU activation.
Projected features are then fused with the main task base features, $f_i$, via an element-wise addition denoted $\bigoplus$, leading to the final refined features $\tilde{f}_i$:
\begin{equation}
    \setlength{\abovedisplayskip}{3pt}
    \setlength{\belowdisplayskip}{3pt}
 	\tilde{f}_{i} = f_i \bigoplus
 	F_{\dir{}{i}} \left( [f_{\dir{j}{i}}|_{j\in T \setminus \{i\}}] \right)\,,
 	\label{eq:base_fusion}
\end{equation}
where $[...]$ is the channel-wise concatenation operation.
Intuitively, the xTAM \textit{directional features} are combined in $F_{\dir{}{i}}\,$ to constitute multi-modal residuals that complement original features $f_i$.
By design, the refined features have same dimensions than the input task-specific features.

Although it is possible to insert a \mainblockshort{} at several levels in the decoder. We show experimentally that the overall best choice is to have a single block before the last layer.
 
\subsubsection{Training.}
To train our framework, we use a multi-scale multi-task objective defined as a linear combination of the task losses %
applied at two or more stages of the decoder: before any \mainblockshort{}, and on the full resolution output. 
Although we do not explicitly introduce losses to encourage cross-task distillation,  this is made possible implicitly by the design of xTAM and \mainblockshort{}.
More details in the supplementary.%

\section{Experiments}
\label{sec:exp}

\corr{We present results} and ablation studies of our proposed framework. 
\corr{While the dense multi-task learning literature lacks a unified setup for benchmarking, we strive to be comprehensive in our evaluation by including 4 tasks combined into 3 tasks sets, and evaluate on 4 datasets with 4~baselines -- each requiring individual retraining.} \cref{sec:exp-setup} describes our experimental and multi-task setup. 

\corr{The evaluation of our contributions is threefold.
First, in \cref{sec:exp_mtl} our MTL proposal is evaluated on three task sets: \SD{} (segmentation + depth), \SDN{} (\textbf{+} normals), and \SDNE{} (\textbf{+} edges).
In \cref{sec:exp_mtl_ablation}, we ablate the position and choice of fusion in \mainblockshort{} and alternative designs for the correlation-guided attention in xTAM.}

\corr{Second, in \cref{sec:exp_seg} we demonstrate the ability of our multi-task strategy to maximize semantic segmentation performance when learning self-supervised depth regression.}

\corr{Third, in \cref{sec:exp_mtlda} we further extend and evaluate our framework for MTL unsupervised domain adaptation.}

\subsection{Experimental setup}
\label{sec:exp-setup}

\subsubsection{Common setup.}
\sloppy
\paragr{Datasets.} 
We leverage two synthetic and two real datasets: \vk~\cite{gaidon2016virtual}, \sy~\cite{ros2016synthia}, \cs~\cite{cordts2016cityscapes}, and  \nyu~\cite{Silberman:ECCV12}.
In Cityscapes, we use depth from stereo via semi-global matching.
Surface normals labels are estimated from depth maps following~\cite{yang2017unsupervised}.
Lastly, \nyu~provides ground-truth semantic edge maps which we use for supervision. We train semantic segmentation on \textit{14} classes on \vk{}, \textit{40} classes on \nyu{}, and \textit{16} classes on \sy{}~and \cs{}.
In VKITTI2 and Synthia, we perform a random 80/20 split.
On \cs~and \nyu, we use the official splits and load the images at $1024{\times}512$ and $576{\times}448$, respectively. They are scaled to $1024{\times}320$ on \vk~and $1024{\times}512$ on \sy.

\paragr{Baselines.}
\corr{We follow the recent survey~\cite{Vandenhende2021} and} compare our performance against single task learning networks~({STL}), where each task is predicted separately by a dedicated networks, and a naive multi-task learning baseline ({MTL}) using a shared encoder and task-specific decoders.
PAD-Net~\cite{xu2018padnet} is a more competitive baseline than MTL.
We address two variants of PAD-Net: (i) the original model used in~\cite{xu2018padnet} and (ii)~a stronger model introduced in~\cite{hoyer2021ways} (`\threewaysPADNet{}'). 
\corr{Seeking fair comparison, note that we applied a substantial experimental effort by best retraining and adapting each baseline to our setups. Details are in the supp.}

\paragr{Architecture.} All networks have a shared ResNet-101~\cite{he2016deep} encoder initialized with ImageNet \cite{imagenet}, and a suitable number of decoders depending on the task set.
\corr{Decoders for STL, MTL and PAD-Net baselines are from the survey \cite{Vandenhende2021}}. While we stick to ~\cite{hoyer2021ways} for \threewaysPADNet{} and our model, using an \textit{Atrous Spatial Pyramid Pooling}~\cite{chen2017deeplab} followed by four upsampling blocks with skip connections~\cite{ronneberger2015u}.

\paragr{Training details.}
We train all models for 40k iterations, set gradient norm clipping to 10 and reduce the learning rates by $0.1$ at 30k steps as in \cite{hoyer2021ways}. 
We train our MTL models with the Adam optimizer~\cite{kingma2014adam} with $\beta_1=0.9$ and $\beta_2=0.98$; learning rate is set to $2.0e{-4}$ for the backbone, and $3.0e{-4}$ for all decoders.
In other models, we use the SGD optimizer, setting the learning rates of the backbone and decoders to $1.0e{-3}$ and $1.0e{-2}$, respectively, as in \cite{hoyer2021ways}. We use momentum of $0.9$ and weight decay of $5.0e{-4}$.

\noindent{}We report the mean and standard deviation over three runs.

\paragr{Supervision.}
We use the cross-entropy for semantic segmentation $\mathcal{L}_S$ and edge estimation $\mathcal{L}_E$~\cite{bruggemann2021exploring}.
The berHu loss \cite{laina2016deeper} is applied on inverse normalized values for depth regression as $\mathcal{L}_D(\hat{d}) = \text{berHu}(d_{far}/\hat{d},\, d_{far}/d)$. 
To supervise surface normals training, we use the approach of \cite{guizilini2021geometric}: $\mathcal{L}_N(\hat{n}) = 1 - \cos(\hat{n},\,n)$, which computes the cosine between the estimated $\hat{n}$ and ground-truth $n$ normal vectors as $\cos({\hat{n},\, n}) = \hat{n}\cdot{}n/\lVert \hat{n} \rVert \lVert n \rVert$.
Losses are weighted using our task balancing strategy.

\paragr{Single-task metrics.}
We report a standard metric for each task: the mean intersection over union (mIoU) for semantic segmentation, the root mean square error in meters (RMSE) for depth regression, the mean error in degrees~(mErr) for surface normals estimation, and the F1-score (F1) for semantic edge estimation.

\subsubsection{Multi-task setup.}
\label{sec:tasks}
\paragr{Task balancing.}
We observe that tasks loss weighting heavily impacts the performance of multi-task models similarly to \cite{Vandenhende2021}, though we also notice that weights vary primarily with the loss functions and their interaction, and much less on the data or model. This is fairly reasonable knowing gradients scale with the loss term magnitude.
Hence, as in~\cite{Vandenhende2021} we applied grid-search for each unique task set, and use the same optimal set of weight for all methods. 
For $T{=}\{\text{S}, \text{D}\}$ we obtain $\{w_S{=}50, w_D{=}1\}$, for $T{=}\{\text{S}, \text{D},\allowbreak \text{N}\}$ we get $\{w_S {=} 100,\allowbreak w_D {=} 1,\allowbreak w_N {=} 100\}$, and choose for $T{=}\{\text{S}, \text{D}, \text{N}, \text{E}\}$, $\{w_S {=} 100,\allowbreak w_D {=} 1,\allowbreak w_N {=} 100,\allowbreak w_E {=} 50\}$. Importantly, this search is done on the MTL baseline ensuring we do not favor our method in any way.

\paragr{MTL metrics.}
Again, our goal is to improve the \textit{overall} performance of the model over a set of tasks $T$, which we measure using the delta metric of~\cite{Vandenhende2021} written $\Delta_{\text{T}}$. 
The latter measures the relative performance of a given multi-task model compared to the performance of dedicated Single Task Learning~(STL) models:
$\Delta_{\text{T}}(\textbf{\text{f}}) = 1/n \sum_{i\in T} (-1)^{g_i}(m_i - b_i)/b_i$, where $m_i$ and $b_i$ are the metrics on task $i$ of the model $\textbf{\text{f}}$ and STL model, respectively, and  $g_i$ is the metric direction of improvement, \ie $g_i{=}0$ when higher is better, and $g_i{=}1$ when lower is better.

\subsection{Main results}
\label{sec:exp_mtl}
\cref{tab:main_mtl} (outdoor) and \cref{tab:nyu} (indoor)  report our main experiments on 4 datasets, and 3 multi-task setups: \SD{}, \SDN{} and \SDNE{}. In almost all setups our cross-task attention mechanism outperforms the baselines.

\corr{From \cref{tab:main_mtl}, on the synthetic Synthia data} with perfect segmentation and depth labels, our model outperforms all baselines by large margins of up to $+5.69 \Delta_\text{SD}$ in \SD{} and up to $+5.75 \Delta_\text{SDN}$ in \SDN{}.
In real Cityscapes data, having stereo depth pseudo labels, our model is outperformed by {\threewaysPADNet{}}, only in the \SD{} setup, and outperforms {\threewaysPADNet{}} in \SDN{}.
The improvements of PAD-Net over MTL baselines confirm the benefit of self-attention mechanism~\cite{xu2018padnet}.
Revisiting the ``3-ways'' architecture introduced in~\cite{hoyer2021ways} in a different context of multi-task learning, we observe a significant leap in performance of \threewaysPADNet{} over PAD-Net; the two models are only different in decoder design.
This shows the importance of decoder in MTL.

\begin{table*}[ht!]
    \captionsetup{font=footnotesize}
	\centering
	\scriptsize
	\begin{tabular}{c l ? c c | c ? c c | c || c | c }
		\toprule
		& \multirow{4}{*}{Methods} & \multicolumn{3}{c?}{\SD{}} & \multicolumn{5}{c}{\SDN{}}     \\
		
		&& \hsem{}Semseg \uarr & \hdepth{}Depth \darr & \hdelta{}Delta \uarr & \hsem{}Semseg \uarr & \hdepth{}Depth \darr & \hdelta{}Delta \uarr & \hnormal{}Normals \darr & \hdelta{}Delta \uarr \\ 
		
		&& \hsem{}\tiny{mIoU \%} & \hdepth{}\tiny{RMSE m} & \color{ogreen}\hdelta{}\tiny{$\dmtl{SD}$ \%} & \hsem{}\tiny{mIoU \%} & \hdepth{}\tiny{RMSE m} & \color{ogreen}\hdelta{}\tiny{$\dmtl{SD}$ \%} & \hnormal{}\tiny{mErr. °} & \hdelta{}\color{purple}\tiny{$\dmtl{SDN}$ \%} \\ 
		
		\midrule
		\multirow{5}{*}{\STAB{\rotatebox[origin=c]{90}{\scriptsize{\sy}}}}
		& STL \cite{Vandenhende2021}    & \perf{67.43}{0.15}  & \perf{5.379}{0.055}  & \reflectbox{\rotatebox[origin=c]{270}{$\drsh$}}  & \textit{\tiny idem} & \textit{\tiny idem}  & \reflectbox{\rotatebox[origin=c]{270}{$\drsh$}} & \perf{19.61}{0.12}  & \reflectbox{\rotatebox[origin=c]{270}{$\drsh$}}  \\ 
		& MTL \cite{Vandenhende2021}    & \perf{69.83}{0.25}  & \perf{5.166}{0.063}  & \color{ogreen}\perf{+03.76}{0.77}  & \perf{71.27}{0.21} & \perf{5.108}{0.076}  & \color{ogreen}\perf{+05.37}{0.83} & \perf{18.51}{0.10}  & \color{purple}\perf{+05.45}{0.72}  \\
		& PAD-Net \cite{xu2018padnet}   & \perf{70.87}{0.15}  & \perf{4.917}{0.014}  & \color{ogreen}\perf{+06.85}{0.24}  & \perf{72.27}{0.25} & \perf{4.949}{0.072}  & \color{ogreen}\perf{+07.58}{0.56} & \perf{19.28}{0.09}  & \color{purple}\perf{+05.62}{0.43}  \\
		& \threewaysPADNet{} \cite{hoyer2021ways}   & \uperf{77.50}{0.17} & \uperf{4.289}{0.028} & \color{ogreen}\uperf{+17.60}{0.13} & \uperf{79.93}{0.5} & \uperf{4.218}{0.082} & \color{ogreen}\uperf{+20.06}{0.92} & \uperf{15.54}{0.14} & \color{purple}\uperf{+20.29}{0.84} \\
		& \oc{}Ours & \oc{}\bperf{80.53}{0.43} & \oc{}\bperf{4.161}{0.022} & \color{ogreen}\oc{}\bperf{+21.04}{0.52} & \oc{}\bperf{82.99}{0.38} & \oc{}\bperf{4.056}{0.076} & \color{ogreen}\oc{}\bperf{+23.83}{0.98} & \oc{}\bperf{14.30}{0.15} & \oc{}\color{purple}\bperf{+24.92}{0.87} \\

		\midrule
		\multirow{5}{*}{\STAB{\rotatebox[origin=c]{90}{\scriptsize{\vk}}}}
		& STL \cite{Vandenhende2021}    & \perf{84.53}{0.06}  & \perf{5.720}{0.027}  & \reflectbox{\rotatebox[origin=c]{270}{$\drsh$}}  & \textit{\tiny idem} & \textit{\tiny idem}  & \reflectbox{\rotatebox[origin=c]{270}{$\drsh$}} & \perf{23.14}{0.68}  & \reflectbox{\rotatebox[origin=c]{270}{$\drsh$}}  \\ 
		& MTL \cite{Vandenhende2021}    & \perf{87.73}{0.12}  & \perf{5.720}{0.029}  & \color{ogreen}\perf{+01.89}{0.21}  & \perf{87.83}{0.21} & \perf{5.714}{0.033} & \color{ogreen}\perf{+02.00}{0.27} & \perf{22.30}{0.68}  & \color{purple}\perf{+02.54}{0.80}  \\
		& PAD-Net \cite{xu2018padnet}   & \perf{88.43}{0.12}  & \perf{5.571}{0.058}  & \color{ogreen}\perf{+03.63}{0.45}  & \perf{88.67}{0.15} & \perf{5.543}{0.043} & \color{ogreen}\perf{+04.09}{0.29} & \perf{22.16}{0.70}  & \color{purple}\perf{+04.09}{0.83}  \\
		& \threewaysPADNet{} \cite{hoyer2021ways}   & \uperf{96.13}{0.15} & \uperf{4.013}{0.051} & \color{ogreen}\uperf{+21.78}{0.54} & \uperf{96.87}{0.06}& \uperf{3.756}{0.013}& \color{ogreen}\uperf{+24.46}{0.14} & \uperf{15.54}{0.56} & \color{purple}\uperf{+27.25}{0.90} \\

		& \oc{}Ours & \oc{}\bperf{97.00}{0.10} & \oc{}\bperf{3.423}{0.025} & \color{ogreen}\oc{}\bperf{+27.47}{0.16} & \oc{}\bperf{97.53}{0.06} & \oc{}\bperf{3.089}{0.006} & \color{ogreen}\oc{}\bperf{+30.70}{0.05} & \oc{}\bperf{14.44}{0.52} & \oc{}\color{purple}\bperf{+33.00}{0.73} \\
		
		\midrule
		\multirow{5}{*}{\STAB{\rotatebox[origin=c]{90}{\scriptsize{Cityscapes}}}}
		& STL \cite{Vandenhende2021}    & \perf{67.93}{0.06}  & \perf{6.622}{0.020}  & \reflectbox{\rotatebox[origin=c]{270}{$\drsh$}}  & \textit{\tiny idem} & \textit{\tiny idem}  & \reflectbox{\rotatebox[origin=c]{270}{$\drsh$}} & \perf{44.10}{0.01}  & \reflectbox{\rotatebox[origin=c]{270}{$\drsh$}}  \\ 
		& MTL \cite{Vandenhende2021}    & \perf{70.43}{0.12}  & \perf{6.797}{0.520}  & \color{ogreen}\perf{+00.52}{0.32}  & \perf{70.93}{0.15} & \perf{6.736}{0.023} & \color{ogreen}\perf{+01.34}{0.28} & \perf{43.60}{0.01}  & \color{purple}\perf{+01.30}{0.18}  \\
		& PAD-Net \cite{xu2018padnet}   & \perf{70.23}{0.25}  & \perf{6.777}{0.010}  & \color{ogreen}\perf{+00.52}{0.27}  & \perf{70.67}{0.06} & \perf{6.755}{0.018} & \color{ogreen}\perf{+01.00}{0.17} & \perf{43.52}{0.00}  & \color{purple}\perf{+01.12}{0.11}  \\
		& \threewaysPADNet{} \cite{hoyer2021ways}   & \bperf{75.00}{0.10} & \bperf{6.528}{0.063} & \color{ogreen}\bperf{+05.91}{0.44} & \uperf{75.50}{0.10}& \uperf{6.491}{0.081}& \color{ogreen}\uperf{+06.56}{0.61} & \uperf{41.84}{0.05} & \color{purple}\uperf{+06.09}{0.37} \\
		
		& \oc{}Ours & \oc{}\uperf{74.95}{0.10} & \oc{}\uperf{6.649}{0.003} & \color{ogreen}\oc{}\uperf{+04.96}{0.08} & \oc{}\bperf{76.08}{0.14} & \oc{}\bperf{6.407}{0.013} & \color{ogreen}\oc{}\bperf{+07.61}{0.04} & \oc{}\bperf{40.05}{0.33} & \oc{}\color{purple}\bperf{+08.15}{0.22} \\

		\bottomrule		
	\end{tabular}
	\caption{\textbf{MTL performance on two sets.} We report individual task metrics but seek best overall performance measured by $\Delta_\text{SD}$ and $\Delta_\text{SDN}$, computed \wrt ``STL''. Except for \SD{} on Cityscapes where we are second, we outperform significantly the baselines on all delta metrics. We also report $\color{ogreen}\Delta_\text{SD}$ in \SDN{} for direct comparison with \SD{}. Notice $\color{ogreen}\Delta_\text{SD}\color{black}({\text{f}}_\text{SDN})>\color{ogreen}\Delta_\text{SD}\color{black}({\text{f}}_\text{SD})$ across methods highlighting the importance of surface normals estimation. We highlight \textbf{best} and \underline{2nd best}.}
	\label{tab:main_mtl}
\end{table*}%

\begin{table*}[ht!]
\captionsetup{font=footnotesize}
	\centering
    \newcommand{\htsem}{Semseg\,$\uparrow$}         %
    \newcommand{\htdepth}{Depth\,$\downarrow$}      %
    \newcommand{\htdelta}{Delta\,$\uparrow$}        %
    \newcommand{\htnormal}{Normals\,$\downarrow$}   %
    \newcommand{\htedge}{Edges\,$\uparrow$}         %
	\resizebox{\linewidth}{!}{%
	\setlength{\tabcolsep}{0.007\linewidth}
	\begin{tabular}{l ? c c | c ? c c || c | c | c ? c c c || c | c | c}

	    \toprule
		&
		\multicolumn{3}{c?}{\SD{}} & 
		\multicolumn{5}{c?}{\SDN{}} & 
		\multicolumn{6}{c}{\SDNE{}} \\
		
		Methods &
		\hsem{}\htsem & \hdepth{}\htdepth & \hdelta{}\htdelta & 
    	\hsem{}\htsem & \hdepth{}\htdepth & \hdelta{}\htdelta & \hnormal{}\htnormal & \hdelta{}\htdelta & 
    	\hsem{}\htsem & \hdepth{}\htdepth & \hnormal{}\htnormal & \hdelta{}\htdelta & \hedge{}\htedge & \hdelta{}\htdelta \\ 
		
		& \hsem{}{\scriptsize{mIoU \%}} & \hdepth{}\scriptsize{RMSE m} & \color{ogreen}\hdelta{}\scriptsize{$\dmtl{SD}$ \%} & 
		\hsem{}\scriptsize{mIoU \%} & \hdepth{}\scriptsize{RMSE m} & \color{ogreen}\hdelta{}\scriptsize{$\dmtl{SD}$ \%} & \hnormal{}\scriptsize{mErr. °} & \color{purple}\hdelta{}\scriptsize{$\dmtl{SDN}$ \%} & 
		\hsem{}\scriptsize{mIoU \%} & \hdepth{}\scriptsize{RMSE m} & \hnormal{}\scriptsize{mErr. °} & \color{purple}\hdelta{}\scriptsize{$\dmtl{SDN}$ \%} & \hedge{}\scriptsize{F1 \%} & \hdelta{}\scriptsize{$\dmtl{SDNE}$ \%} \\ 
		
		\midrule
		
		STL \cite{Vandenhende2021} & \perf{38.70}{0.10}  & \perf{0.635}{0.013}  & \reflectbox{\rotatebox[origin=c]{270}{$\drsh$}}  & \textit{\footnotesize idem} & \textit{\footnotesize idem} & \reflectbox{\rotatebox[origin=c]{270}{$\drsh$}} & \perf{36.90}{0.26} & \reflectbox{\rotatebox[origin=c]{270}{$\drsh$}} & \textit{\footnotesize idem} & \textit{\footnotesize idem} & \textit{\footnotesize idem} & \reflectbox{\rotatebox[origin=c]{270}{$\drsh$}} & \perf{54.90}{0.00} & \reflectbox{\rotatebox[origin=c]{270}{$\drsh$}}  \\ 
		
        MTL \cite{Vandenhende2021} & \uperf{39.44}{0.34} & \perf{0.638}{0.004} & \color{ogreen}\perf{+1.63}{0.37} & \uperf{39.90}{0.41} & \perf{0.642}{0.003} & \color{ogreen}\perf{+1.89}{0.67} & \perf{36.07}{0.09} & \color{purple}\perf{+1.76}{0.53} & \perf{39.70}{0.35} & \perf{0.636}{0.001} & \perf{36.10}{0.12} & \color{purple}\perf{+1.88}{0.33} & \perf{55.11}{0.15} & \perf{+1.50}{0.20} \\

		PAD-Net \cite{xu2018padnet} & \perf{35.30}{0.84} & \perf{0.659}{0.004} & \color{ogreen}\perf{-5.36}{0.83} & \perf{36.14}{0.30} & \perf{0.660}{0.006} & \color{ogreen}\perf{-4.32}{0.68} & \perf{36.72}{0.08} & \color{purple}\perf{-2.97}{0.43} & \perf{36.19}{0.24} & \perf{0.662}{0.005} & \perf{36.58}{0.06} & \color{purple}\perf{-2.92}{0.37} & \perf{54.79}{0.07} & \perf{-2.24}{0.26} \\
		
		\threewaysPADNet{} \cite{hoyer2021ways} & \bperf{39.47}{0.16} & \uperf{0.622}{0.001} & \color{ogreen}\uperf{+2.90}{0.23} & \bperf{40.28}{0.30} & \uperf{0.619}{0.004} & \color{ogreen}\uperf{+4.16}{0.50} & \uperf{35.35}{0.09} & \color{purple}\uperf{+3.93}{0.27} & \uperf{40.16}{0.28} & \uperf{0.614}{0.010} & \uperf{35.25}{0.09} & \color{purple}\uperf{+4.14}{0.65} & \uperf{59.66}{0.16} & \uperf{+5.27}{0.49} \\
		
		\oc{}Ours & \oc{}\perf{38.93}{0.35} & \oc{}\bperf{0.604}{0.004} & \color{ogreen}\oc{}\bperf{+3.54}{0.21} & \oc{}\bperf{40.28}{0.41} & \oc{}\bperf{0.598}{0.002} & \color{ogreen}\oc{}\bperf{+5.80}{0.65} & \oc{}\bperf{33.72}{0.14} & \color{purple}\oc{}\bperf{+6.49}{0.50} & \oc{}\bperf{40.84}{0.37} & \oc{}\bperf{0.593}{0.004} & \oc{}\bperf{33.38}{0.19} & \color{purple}\oc{}\bperf{+7.52}{0.27} & \oc{}\bperf{61.12}{0.24} & \oc{}\bperf{+8.47}{0.12} \\
		
		\bottomrule		
	\end{tabular}}
    \caption{\textbf{Results on \nyu{}~ evaluated on three sets of tasks.} We report a delta metric for each set as in \cref{tab:main_mtl} and provide partial metrics $\color{ogreen}\dmtl{SD}$$(\text{f}_{\text{SDN}})$ and $\color{purple}\dmtl{SDN}$$(\text{f}_{\text{SDNE}})$ to compare between sets and highlight the benefit of inserting additional tasks in the framework. We underline the fact that \mbox{$\color{purple}\Delta_\text{SDN}$$({\text{f}}_\text{SDNE})>\color{purple} \Delta_\text{SDN}$$({\text{f}}_\text{SDN})$} validating the benefit of edge estimation even for the other tasks.} 
	\label{tab:nyu}
\end{table*}

\begin{figure*}[!t]
\centering
\captionsetup{font=footnotesize}
\setlength{\tabcolsep}{0.0ex}
\resizebox{0.98\textwidth}{!}{
    \begin{tabular}{c c c c c} %
    	\hdelta{}Input&{\color{white}abc}&\hsem{}Segmentation&\hdepth{}Depth&\hnormal{}Normals\\
    	
    	\includegraphics[width=0.26\linewidth,valign=m]{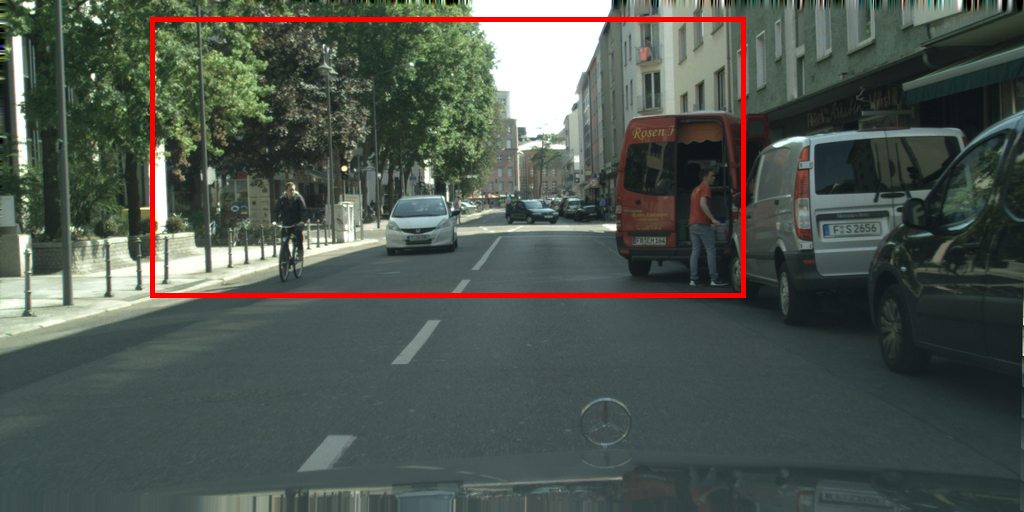}
    	&\STAB{\rotatebox[origin=c]{90}{\footnotesize{ground-truth}}}
    	&\includegraphics[width=0.26\linewidth,valign=m]{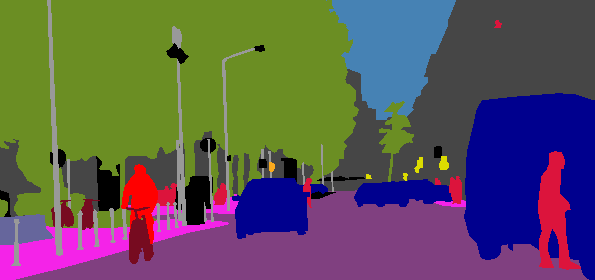}
    	&\includegraphics[width=0.26\linewidth,valign=m]{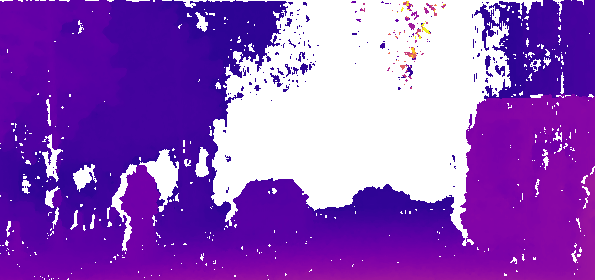}
    	&\includegraphics[width=0.26\linewidth,valign=m]{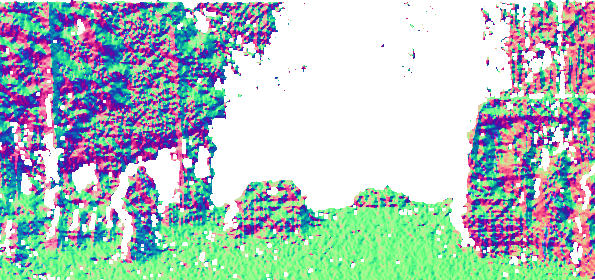}\\
    	\cmidrule{3-5}
    	
    	&\STAB{\rotatebox[origin=c]{90}{\footnotesize{PAD-Net~\cite{xu2018padnet}}}}
    	&\includegraphics[width=0.26\linewidth,valign=m]{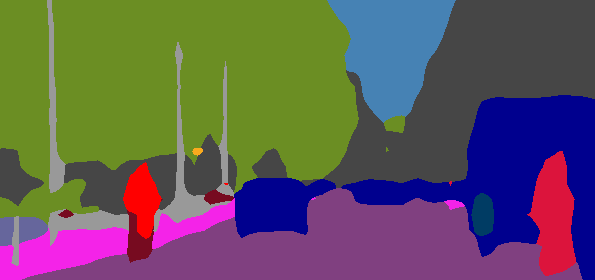}
    	&\includegraphics[width=0.26\linewidth,valign=m]{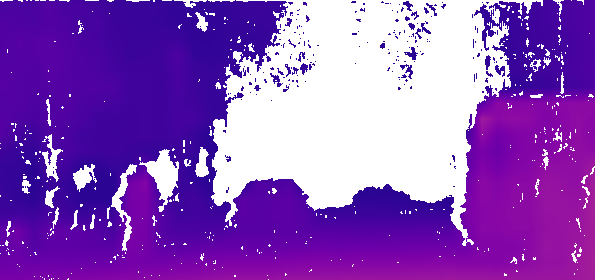}
    	&\includegraphics[width=0.26\linewidth,valign=m]{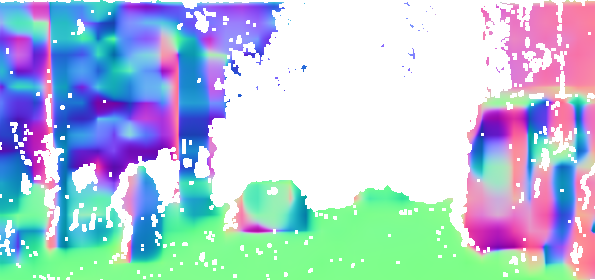}\\
    	
    	&\STAB{\rotatebox[origin=c]{90}{\footnotesize{\threewaysPADNet{}~\cite{hoyer2021ways}}}}
    	&\includegraphics[width=0.26\linewidth,valign=m]{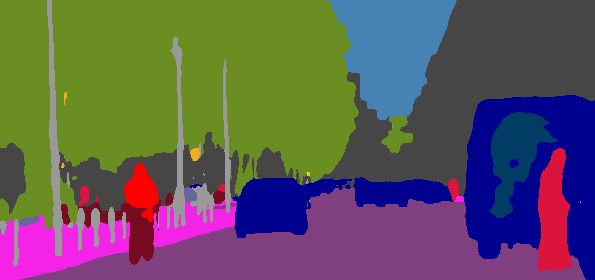}
    	&\includegraphics[width=0.26\linewidth,valign=m]{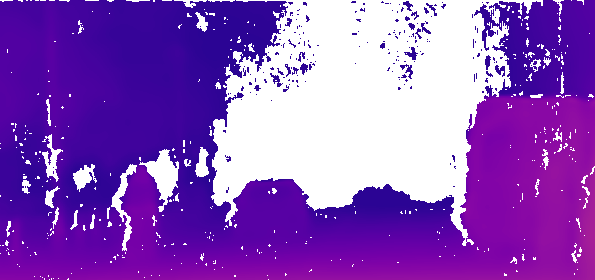}
    	&\includegraphics[width=0.26\linewidth,valign=m]{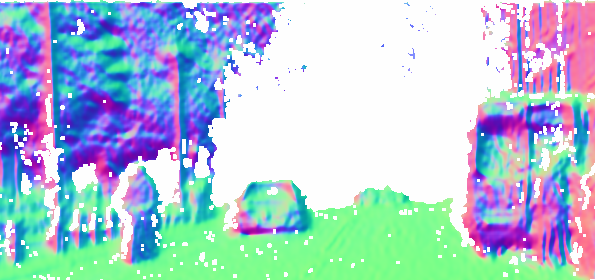}\\
    	
    	&\STAB{\rotatebox[origin=c]{90}{\footnotesize{Ours}}}
    	&\includegraphics[width=0.26\linewidth,valign=m]{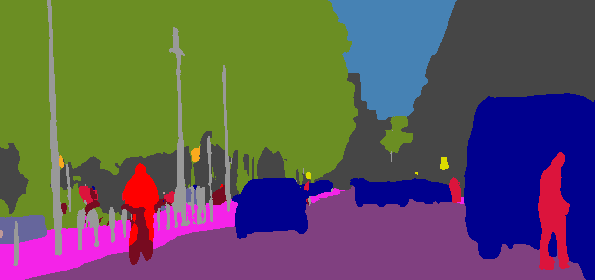}
    	&\includegraphics[width=0.26\linewidth,valign=m]{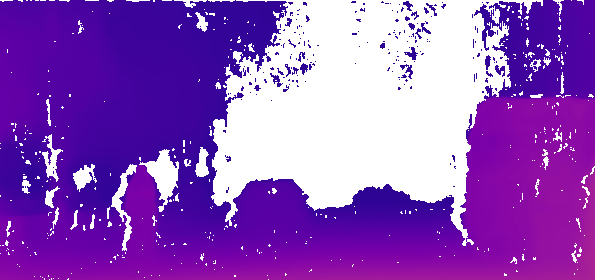}
    	&\includegraphics[width=0.26\linewidth,valign=m]{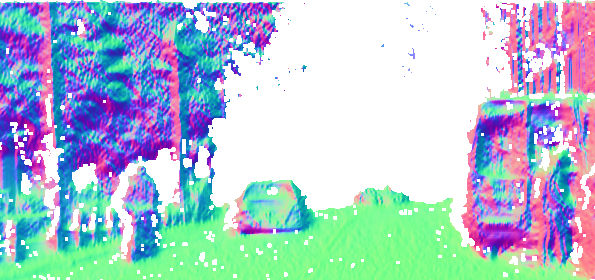}\\
    	
    \end{tabular}%
}
\caption{\textbf{Qualitative MTL results on \qualdata{} in the \SDN{} setup.} The first row shows the input image and its segmentation, depth and normal ground-truths. Segmentation results of our models are better overall, especially in boundary areas -- see ``bicycle'', ``rider'' and ``pedestrian''. Surface normals of PAD-Net is blurrier as compared to \threewaysPADNet{} and Ours.}
\label{fig:mtl_quali_small}
\end{figure*}

\noindent{}Furthermore, the best overall performance (\ie delta metrics) coincides with the best individual task performances, which shows the benefit of handling multiple tasks together.
It is striking to look at the \SDN{} setup, where we also account for the rarely studied normal estimation task. 
In addition to the $\Delta_\text{SDN}({\text{f}}_\text{SDN})$ metric, we report a partial delta measurement $\Delta_\text{SD}({\text{f}}_\text{SDN})$. This allows us to compare ${\text{f}}_\text{SD}$ to ${\text{f}}_\text{SDN}$, respectively, the \SD{} and \SDN{} models considering normal estimation as an auxiliary task. We record a gap of up to +3.23 points between $\Delta_\text{SD}({\text{f}}_\text{SDN})$ and $\Delta_\text{SD}({\text{f}}_\text{SD})$, showing the benefit of injecting additional geometrical cues in the form of surface normals to help the other two tasks.
We note that Ours follows this same observation, having $\Delta_\text{SD}({\text{f}}_\text{SD})<\Delta_\text{SD}({\text{f}}_\text{SDN})$ across all datasets.

In \cref{tab:nyu} we report indoor results on \nyu{}, which has additional Edge labels, and thus reporting \SD{}, \SDN{}, and the newly formed \SDNE{}. Our method outperforms all baselines in the 3 task sets. The partial delta metrics $\Delta_{\text {SDN}}$$(\text{f}_{\text{SD}})$ and $\Delta_{\text{SDN}}$$(\text{f}_{\text{SDNE}})$ underline the advantage of normals and edges estimation, respectively. Using edge information for `Ours' not only improved the partial $\Delta_{\text{SDN}}$ in \SDNE{} compared to \SDN{} (\ie, +7.52 vs +6.49) but also led to better individual S/D/N metrics. This shows our framework benefits from additional cues for better scene understanding.

Figure~\ref{fig:mtl_quali_small} shows qualitative results on the \qualdata{} dataset. This is noticeable when looking at thin elements (\eg pedestrians, bicycles, \textit{etc.}) and object contours. The visuals are in fair agreement with the quantitative analysis.

\subsection{Ablation studies}
\label{sec:exp_mtl_ablation}
We report ablations of our cross-task distillation module on the VKITTI2 dataset on both the \SD{} and \SDN{} setups. We evaluate our method using different combinations of our module, varying the position and number of exchange blocks (\cref{tab:ablate_mtl}), changing the type of fusion used (\cref{tab:ablate_fusion}), or the type of attention used (\cref{tab:ablate_attention}). 

\paragr{Multi-task exchange.} In \cref{tab:ablate_mtl} we report variations of our MTL framework with \mainblockshort{} (cf.~\cref{sec:meth-mtl-refblock}) positioned at different scales in the decoders, where scale $s$ is the number of downsamplings from input resolution. We observe that with a single block (upper part, \cref{tab:ablate_mtl}), the $\dmtl{SD}$ and $\dmtl{SDN}$ metrics are lower if the block is placed early in the decoder (\ie scale 4) as opposed to being later in the decoder (\ie scale 1). This follows the intuition that late features are more task-specific, thus better appropriate for cross-task distillation.
Using several blocks (lower part, \cref{tab:ablate_mtl}) was shown marginally beneficial, and at a cost of significantly more parameters (`Param.' column). 

\paragr{Multi-task features fusion.} Considering a single \mainblockshort{} at scale 1, we replace the original \textit{add} fusion operator in \cref{eq:base_fusion} with either a \textit{concatenation} or a \textit{product} and report results in \cref{tab:ablate_fusion} showing that this choice impacts the overall delta metric. %
We notice that the gap differs on other datasets though `add' and `prod' remain always best.

\paragr{Correlation-guided attention.} While our original xTAM design (cf.~\cref{sec:xtam}) uses spatial attention with \cref{eq:crosstaskji} for cross-task correlation-guided exchange, other practices exist in the literature~\cite{vandenhende2020mtinet}.
Hence, in \cref{tab:ablate_attention} we replace our choice of attention in \cref{eq:crosstaskji} using either \textit{spatial} attention, \textit{channel} attention, or \textit{both}. Results show that \textit{channel} attentions or \textit{both} are less efficient here, and come at higher implementation complexity.

\noindent In the appendix, we show it is preferred to combine correlation-guided attention and self-attention in xTAM.

\begin{table*}%
\centering
\scriptsize
\setlength{\tabcolsep}{0.003\textwidth}
\captionsetup{font=footnotesize}
    \begin{minipage}{0.66\textwidth}
        \begin{tabular}{c c c c | c| c c | c ? c|c c c | c }
    		\toprule
    		\multicolumn{4}{c|}{\mainblockshort{}} & \multicolumn{4}{c?}{\SD{}} & \multicolumn{5}{c}{\SDN{}} \\[0.5ex]
    		
    		\multicolumn{4}{c|}{Scales} & Param. \darr& \hsem{}Semseg \uarr & \hdepth{}Depth \darr & \hdelta{}Delta \uarr & Param. \darr& \hsem{}Semseg \uarr & \hdepth{}Depth \darr & \hnormal{}Normals \darr & \hdelta{}Delta \uarr \\ 
    		
    		4&3&2&1 & \tiny{\#M added} & \hsem{}\tiny{mIoU \%} & \cellcolor{xorange}\tiny{RMSE m} & \cellcolor{xgray}\tiny{$\dmtl{SD}$ \%} & \tiny{\#M added} & \hsem{}\tiny{mIoU \%} & \cellcolor{xorange}\tiny{RMSE m} & \cellcolor{xgreen}\tiny{mErr. °} & \cellcolor{xgray}\tiny{$\dmtl{SDN}$ \%} \\
    
    		\midrule
    		
    		&&&& \param{\textbf{0.00}} & \perf{96.88}{0.30} & \perf{3.604}{0.020} & \perf{25.81}{0.33} & \param{\textbf{0.00}} & \perf{97.38}{0.02} & \perf{3.491}{0.041} & \perf{14.50}{0.57} & \perf{30.51}{0.98} \\
    		
    		\midrule
    		\tick&&&& \param{\underline{3.09}} & \bperf{97.32}{0.06} & \perf{3.556}{0.029} & \perf{26.50}{0.29} & \param{\underline{2.32}} &\perf{97.43}{0.04} & \perf{3.559}{0.024} & \perf{14.51}{0.50} & \perf{30.12}{0.79} \\
    		&\tick&&& \param{\underline{3.09}} &\uperf{97.24}{0.03} & \perf{3.476}{0.018} & \uperf{27.15}{0.16} & \param{\underline{2.32}} &\uperf{97.49}{0.08} & \perf{3.353}{0.025} & \perf{14.45}{0.48} & \perf{31.43}{0.76} \\
    		&&\tick&& \param{\textbf{0.77}} &\perf{97.07}{0.06} & \uperf{3.468}{0.016} & \perf{27.11}{0.12} & \param{\textbf{9.26}} &\perf{97.47}{0.06} & \uperf{3.244}{0.035} & \uperf{14.57}{0.51} & \uperf{31.89}{0.92} \\
    		&&&\tick& \param{\textbf{0.77}} &\perf{97.00}{0.10} & \bperf{3.423}{0.025} & \bperf{27.47}{0.16} & \param{\textbf{9.26}} &\bperf{97.53}{0.06} & \bperf{3.089}{0.006} & \bperf{14.44}{0.52} & \bperf{33.00}{0.73} \\
    		
    		\midrule
    		&\tick&&\tick& \param{\textbf{3.86}} &\bperf{97.09}{0.03} & \bperf{3.369}{0.022} & \bperf{27.99}{0.18} & \param{\textbf{11.58}} &\bperf{97.53}{0.02} & \bperf{3.080}{0.025} & \bperf{14.47}{0.57} & \bperf{33.02}{0.90} \\
    		&\tick&\tick&\tick& \param{\underline{4.63}} &\perf{97.01}{0.02} & \uperf{3.377}{0.008} & \perf{27.88}{0.06} & \param{\underline{13.89}} &\uperf{97.39}{0.02} & \uperf{3.136}{0.046} & \uperf{14.81}{0.77} & \uperf{32.13}{1.39} \\
    		\tick&\tick&\tick&\tick& \param{7.72} &\uperf{97.05}{0.03} & \bperf{3.369}{0.010} & \uperf{27.97}{0.08} & \param{23.15} &\perf{96.82}{0.23} & \perf{3.307}{0.066} & \perf{15.39}{0.65} & \perf{30.08}{1.39} \\
    		\bottomrule
    	\end{tabular}%
        
        \caption{\textbf{Ablations on \vk{} changing placement of \mainblockshort{}.} The upper part shows our block is more efficient when located later in the decoders (\ie scale 1) which we attribute to more task-specific features. Using more blocks leads to a small boost at the cost of more parameters (`Param.' column represents the size of distillation block).}%
        \label{tab:ablate_mtl}%
    \end{minipage}%
    \hfill
    \begin{minipage}{0.33\textwidth}
        \subfloat[][Fusion operation]{
            \begin{tabular}{l | c c c | c }
        		\toprule
        		& \multicolumn{4}{c}{\SDN{}} \\[0.5ex]
        			
        		Ablation & \hsem{}Semseg \uarr & \hdepth{}Depth \darr & \hnormal{}Normals \darr & \hdelta{}Delta \uarr \\ 
        		& \hsem{}\tiny{mIoU \%} & \hdepth{}\tiny{RMSE m} & \hnormal{}\tiny{mErr. °} & \hdelta{}\tiny{$\dmtl{SDN}$ \%} \\
        			
        		\midrule 
        		concat & \perf{97.43}{0.06} & \perf{3.315}{0.099} & \perf{14.83}{0.43} & \perf{31.08}{0.61} \\
        		prod & \bperf{97.63}{0.06} & \uperf{3.139}{0.024} & \bperf{14.34}{0.49} & \uperf{32.90}{0.81} \\ 
        		add & \uperf{97.53}{0.06} & \bperf{3.089}{0.006} & \uperf{14.44}{0.52} & \bperf{33.00}{0.73} \\
        		\midrule
    		\end{tabular}
            
            \label{tab:ablate_fusion}}\\
        \subfloat[][Type of attention]{
            \begin{tabular}{l | c c c | c }
    			\midrule
    			\multirow{1}{*}{\STAB{{spatial}}} & \bperf{97.43}{0.06} & \bperf{3.315}{0.099} & \bperf{14.83}{0.43} & \bperf{31.08}{0.61} \\
    			\multirow{1}{*}{\STAB{{channel}}} & \perf{97.08}{0.24} & \perf{3.585}{0.047} & \perf{15.03}{0.51} & \perf{28.70}{0.92} \\
    			\multirow{1}{*}{\STAB{{both}}} & \uperf{97.33}{0.11} & \uperf{3.352}{0.068} & \uperf{14.85}{0.48} & \uperf{30.42}{0.51} \\
    			\bottomrule
    		\end{tabular}
            \label{tab:ablate_attention}}      
        
        \caption{\textbf{Ablation on \vk{}.} We compare different fusion operators for \cref{eq:base_fusion}, in \ref{tab:ablate_fusion} and different choices of attention for xTAM, in \ref{tab:ablate_attention}.}
        \label{tab:ablate_modules}%
    \end{minipage}%
\end{table*}

\subsection{MTL for segmentation}
\label{sec:exp_seg}
We now evaluate our multi-task strategy by focusing on semantic segmentation -- often seen as a core task for scene analysis.
In~\cref{tab:cs_ssde}, we report performance training with two tasks (semantics+depth) on `Cityscapes SDE' where unlike prior results we use \textbf{S}elf-supervised monocular \textbf{D}epth \textbf{E}stimation (SDE) with Monodepth2~\cite{godard2019digging}, as in ``3-ways''~\cite{hoyer2021ways}.
In the latter, a proposed strategy is to use the multi-task module of PAD-Net (``\threewaysPADNet{}'' in~\cref{tab:cs_ssde}).
We  proceed similarly but replacing the above module with our \mainblockshort{} (hereafter, ``\threewaysOurs{}''). We refer to~\cite{hoyer2021ways} and the supplementary for in-depth technical details.
Note that neither ``\threewaysPADNet{}'' nor ``\threewaysOurs{}''use the strategies of ``3-ways'' like data augmentation and selection for annotation. Still,~\cref{tab:cs_ssde} shows that using our module, \ie \threewaysOurs{}, achieves best results ($+3.20$ mIoU).
This shows the increased benefit of our multi-task module.

\begin{table*}[ht!]
	\centering
	\captionsetup{font=footnotesize}
	\setlength{\tabcolsep}{0.004\textwidth}
    \resizebox{\linewidth}{!}{%
		\begin{tabular}{lr|cccccccccccccccc|c}
			\toprule
			Methods && 
			{${\color{csroad}\blacksquare}$ road} & 
			{${\color{csside}\blacksquare}$ swalk} & 
			{${\color{csbuild}\blacksquare}$ build.} & 
			{${\color{cswall}\blacksquare}$ wall}& 
			{${\color{csfence}\blacksquare}$ fence}& 
			{${\color{cspole}\blacksquare}$ pole}& 
			{${\color{cslight}\blacksquare}$ light}& 
			{${\color{cssign}\blacksquare}$ sign}&
			{${\color{csveg}\blacksquare}$ veg}&
			{${\color{cssky}\blacksquare}$ sky}&
			{${\color{csperson}\blacksquare}$ person}&
			{${\color{csrider}\blacksquare}$ rider}&
			{${\color{cscar}\blacksquare}$ car}&
			{${\color{csbus}\blacksquare}$ bus}&
			{${\color{csmbike}\blacksquare}$ mbike}&
			{${\color{csbike}\blacksquare}$ bike}&
			{mIoU\,\%}\\
			\midrule
			3-ways & \cite{hoyer2021ways} &\footnotesize{*}&\footnotesize{*}&\footnotesize{*}&\footnotesize{*}&\footnotesize{*}&\footnotesize{*}&\footnotesize{*}&\footnotesize{*}&\footnotesize{*}&\footnotesize{*}&\footnotesize{*}&\footnotesize{*}&\footnotesize{*}&\footnotesize{*}&\footnotesize{*}&\footnotesize{*}& \perfn{71.16}\\
			
			\threewaysPADNet{} & \cite{hoyer2021ways} & \perfn{97.21} &\perfn{79.38} &\perfn{90.50} &\perfn{\textbf{47.68}} &\perfn{49.68} &\perfn{51.17} &\perfn{49.41} &\perfn{64.65} &\perfn{91.40} &\perfn{93.85} &\perfn{72.41} &\perfn{46.92} &\perfn{92.66} &\perfn{80.17} &\perfn{42.43} &\perfn{66.39} & \perfn{69.74} \\
			
			\threewaysOurs{} & ours & \perfn{\textbf{97.62}} &\perfn{\textbf{82.29}} &\perfn{\textbf{92.44}} &\perfn{46.52} &\perfn{\textbf{54.76}} &\perfn{\textbf{59.82}} &\perfn{\textbf{60.94}} &\perfn{\textbf{73.13}} &\perfn{\textbf{92.22}} &\perfn{\textbf{94.55}} &\perfn{\textbf{76.40}} &\perfn{\textbf{58.49}} &\perfn{\textbf{94.26}} &\perfn{\textbf{85.14}} &\perfn{\textbf{49.41}} &\perfn{\textbf{71.70}} & \perfn{\textbf{74.36}} \\
			\bottomrule
		\end{tabular}
		}\\
		
	\caption{\textbf{Semantic segmentation with SDE on \cs{}.} We insert PAD-Net distillation block or \mainblockshort{}: ``\threewaysPADNet{}'' or ``\threewaysOurs{}'', respectively, in the architecture of \cite{hoyer2021ways}. * We only report mIoU because \cite{hoyer2021ways} does not release their model weights trained on the full set.}
\label{tab:cs_ssde}
\vspace{-1em}
\end{table*}

\subsection{MTL for UDA}
\label{sec:exp_mtlda}

Unsupervised domain adaptation (UDA) is the line of research dealing with distribution shifts between the \textit{source} domain, in which we have labeled data, and the \textit{target} domain, in which only unlabeled data is available for training.
Here, we extend our experiments to the novel MTL Unsupervised Domain Adaptation setup (MTL-UDA) where the goal is to perform well \textit{in average} on all tasks in the target domain.
We argue that task exchange is beneficial for MTL-UDA since semantic- and geometry-related tasks exhibit different behaviors and sensitivities to the shift of domain, and are shown to be complementary.

We leverage typical synthetic to real scenarios: \sy$\mapsto$\cs{} and \vk$\mapsto$\cs{}, reporting results on the \SD{} set. While semantic segmentation is degraded due to the shifts in colors, depth estimation is more affected by changes in scene composition and sensors.

\paragr{Architecture adjustments.} We adopt a simple multi-task Domain Adaptation~(DA) solution based on output-level DA adversarial training~\cite{vu2019advent,vu2019dada,saha2021learning}.
As output-level techniques do not alter the base MTL architecture, it allows direct understanding whether our MTL design is favorable or not for adaptation.
For each task, a small discriminator network is jointly trained with the main MTL model, which act as two sides in an adversarial game.
While discriminators try to tell from which domain the input data comes from, the MTL model tries to fool all discriminators by making outputs of source and target domain indistinguishable, which eventually helps align source/target.
The complete network and training are detailed in the supplementary materials.

\paragr{Baselines adjustments.} We use the same baselines as in~\cref{sec:exp_mtl} and introduce patch discriminators to align the processed output prediction maps. %
We use the STL-UDA models - consisting of a single task decoder and one output type - as baseline for measuring the delta metric.
We train all methods on source and target data in mix batches containing two instances from either domains. 

\paragr{Results.} In \cref{tab:mtlda}, results show that the use of our multi-task exchange significantly improves performance in all scenarios and metrics. Along with MTL results, we report STL \textit{source}, trained only on source, and STL \textit{oracle}, trained on labeled target. 
In \sy$\mapsto$\cs{} our method very significantly outperforms the naive MTL-UDA baseline by $+38.1\Delta_{\text{SD}}$, and by $+8.34\Delta_{\text{SD}}$ in \vk$\mapsto$\cs{}, also coinciding with the best individual task metrics.
An interesting note is that STL-UDA is preferable over straightforward multi-task learning in UDA~(\mbox{MTL-UDA}).

\begin{table}{}
	\centering
	\scriptsize
	\captionsetup{font=footnotesize}
	\setlength{\tabcolsep}{0.003\textwidth}%
	\begin{tabular}{l ? c c | c ? c c | c}
	
		\toprule
		& \multicolumn{3}{c?}{\sy{}$\mapsto$\cs{} \tiny{(16 classes)}} & \multicolumn{3}{c}{\vk{}$\mapsto$\cs{} \tiny{(8 classes)}} \\[0.5ex]
		\multirow{2}{*}{Methods} & \hsem{}Semseg \uarr & \hdepth{}Depth \darr & \hdelta{}Delta \uarr & \hsem{}Semseg \uarr & \hdepth{}Depth \darr & \hdelta{}Delta \uarr \\
		
		& \hsem{}\tiny{mIoU \%} & \hdepth{}\tiny{RMSE m} & \hdelta{}\tiny{$\dmtl{SD}$ \%} & \hsem{}\tiny{mIoU \%} & \hdepth{}\tiny{RMSE m} & \hdelta{}\tiny{$\dmtl{SD}$ \%} \\ 
		\midrule
	
		STL \textit{target} & \perf{67.93}{0.06} & \perf{06.62}{0.02} & \large{-} & \perf{77.10}{0.10} & \perf{06.62}{0.02} & \large{-}\\
		STL \textit{source} & \perf{35.63}{0.67} & \perf{13.79}{0.28} & \large{-} & \perf{58.77}{0.06} & \perf{11.99}{0.34} & \large{-}\\
		MTL \textit{target} & \perf{70.43}{0.12}  & \perf{06.79}{0.52} & \large{-} & \perf{79.63}{0.07} & \perf{06.72}{0.02} & \large{-}\\
		\midrule
		STL-UDA & \uperf{37.55}{0.92} & \uperf{14.26}{0.14} & \reflectbox{\rotatebox[origin=c]{270}{$\drsh$}} & \uperf{61.60}{0.35} & \uperf{11.45}{0.32} & \reflectbox{\rotatebox[origin=c]{270}{$\drsh$}} \\
		MTL \textit{source} & \perf{15.32}{1.02} & \perf{14.51}{0.42} & \perf{-30.49}{2.58} & \perf{49.50}{1.64} & \perf{12.26}{0.69} & \perf{-13.37}{2.67}\\
		MTL-UDA & \perf{16.71}{1.22} & \perf{14.47}{0.86} & \perf{-28.49}{2.41} & \perf{57.26}{0.71} & \perf{11.85}{0.32} & \perf{-05.31}{0.82}\\
		\oc{}Ours & \oc{}\bperf{37.93}{0.79} & \oc{}\bperf{11.66}{0.55} & \oc{}\bperf{09.61}{0.87} & \oc{}\bperf{63.76}{0.91} & \oc{}\bperf{11.15}{0.08} & \oc{}\bperf{03.03}{0.92}\\
		\bottomrule
	\end{tabular}
	\caption{\textbf{MTL-UDA.} We consider the \SD{} task set and report $\Delta_\text{SD}$ metric \wrt{} single-task learning unsupervised domain adaptation (STL-UDA). Overall, naive multi-task adaptation strategy (MTL-UDA) is lowering performance \wrt{} STL-UDA, while our method outperforms all others}
	\label{tab:mtlda}
	\vspace{-2em}
\end{table}

\section{Conclusions}
We address semantic and geometry scene understanding with a multi-task learning framework on the four tasks: semantic segmentation, dense depth regression, surface normal estimation, and edge estimation.
We propose a novel distillation module, \mainblock, built upon xTAM, a bidirectional cross-task attention mechanism that combines two types of cross-task attention to refine and enhance all task-specific features.
Extensive experiments in various datasets demonstrate the effectiveness of our proposed framework over competitive MTL baselines.
We extend our work to two scenarios: (i) to improve semantic segmentation on a recent self-supervised framework~\cite{hoyer2021ways} and (ii) to improve the performance of MTL-UDA.

\vspace{2ex}
\noindent\textbf{Acknowledgements.}
This work was funded by the French Agence Nationale de la Recherche (ANR) with project SIGHT (ANR-20-CE23-0016) performed with HPC resources from GENCI-IDRIS (Grant 2021-AD011012808).

\clearpage
\appendix

In this supplementary material, we provide information about the method in \ref{sec:supp-method-training}, experimental details and additional results in \ref{sec:supp-exp}, and finally quantitative results in \ref{sec:supp-results}.

\section{Method details}
\label{sec:supp-method}

We provide details for the training of our method and some ablations of our multi-task exchange block (\mainblockshort{}). %

\subsection{Training}
\label{sec:supp-method-training}
Considering $T=\{T_1, \dots, T_n\}$ the set of $n$ tasks to \corr{be jointly optimized}, our general MTL training loss is a weighted combination of individual tasks losses and writes:
\begin{equation}
	\mathcal{L}_{\text{tasks}} = \frac{1}{\lvert S \rvert}\sum_{s \in S}{\sum_{t \in T}{\omega_{t}\mathcal{L}^s_t}}\, + {\sum_{t \in T}{\omega_{t}\mathcal{L}^\text{final}_t}},
\end{equation}
where $\omega_{t}$ is the task-balancing weight, $\mathcal{L}_t$ is the task-specific supervision loss at intermediate scale $s$ (\ie $\mathcal{L}^s_t$) or at the final prediction stage (\ie $\mathcal{L}^\text{final}_t$). See Sec. {\color{red}4.1.1} of main paper for details on the loss functions used.
Additionally, $S$ defines the intermediate scales of supervision considering scale $s$ to be the task intermediate output at resolution $\frac{1}{2^{{s}}}$ \wrt input resolution.
In practice, to enforce cross-task exchange without direct mTEB supervision, we set $S$ equal to the scales \corr{at which the mTEB are inserted}. In our \corr{method, we keep a single mTEB and set $S=\{1\}$}.

All our models are trained on a single V100 32g GPU and take between 10 and 20 hours to converge depending on the task set and dataset size.

\subsection{Ablations}
\label{sec:supp-method-ablation}
We study the overall benefit of our multi-task exchange block (mTEB) on \vk{} with our complete architecture.
To further demonstrate the effect of attention mechanisms in our method, we remove self-attention from the directional features of xTAM, so that Eq. ({\color{red}4}) of main paper now writes: %
$f_{\dir{j}{i}} = [\text{diag}(\alpha_1,...,\alpha_c) \times \textbf{xtask}_{\boldsymbol{\dir{j}{i}}}]\,$.
We report the performance of~\SDN{} with spatial cross-task attention. \textit{Without} self-attention we get 97.40/3.556/\textbf{14.39}$\Vert$30.30 for mIoU/RMSE/mErr$\Vert\Delta_\text{SDN}$, and \textbf{97.43}/\textbf{3.315}/14.83$\Vert$\textbf{31.08} \textit{with} self-attention. This shows the two types of attention are best combined.

\section{Experimental details}
\label{sec:supp-exp}

\subsection{Multi-task setup}

\subsubsection{Metrics.}
In the following paragraph we detail the four task-specific metrics used throughout our paper.
\begin{itemize}
  \item[$\bullet$] \textbf{Semantic segmentation} uses mIoU as the average of the per-class Intersection over Union (\%) between label $s$ and predicted map $\hat{s}$ : ${m_S = \text{mIoU}(\hat{s}, s)}$.

  \item[$\bullet$] \textbf{Depth \corr{regression}} uses the Root Mean Square Error computed between label $d$ and predicted map $\hat{d}$: $m_D = \text{RMSE}(\hat{d}, d)$, reporting the RMSE in meters over the evaluated set of images. In the SDE and DA setups, a \textit{per}-image median scaling~\cite{zhou2017sde} is applied since the models used are \textit{not} scale-aware. %

  \item[$\bullet$] \textbf{Normals estimation}, we measure the absolute angle error in degrees between the label $n$ and predicted map $\hat{n}$: $m_N = \text{deg}(\hat{n}, n)$. For all datasets we retrieve labels from the depth map~\cite{yang2017unsupervised}: we unproject the pixels using the camera intrinsics and depth values, then compute the cross-product using neighboring points (from 2D perspective) \corr{\cite{guizilini2021geometric}~}and average over pairs of neighbors \cite{Yang_Wang_Xu_Zhao_Nevatia_2018}. \corr{\cs~\cite{cordts2016cityscapes}~provides disparity maps which we use to compute noisy surface normals labels.}

  \item[$\bullet$] \textbf{Edge estimation}, we apply the F1-score between the predicted and ground-truth maps: $m_E = \text{F1}(\hat{e}, e)$. \cite{Vandenhende2021} provides ground-truth semantic edges for \nyu.
\end{itemize}

\subsubsection{Task balancing.}
\cref{tab:mtl-gridsearch} reports a subset of our grid-search to select an optimal set of weights for both \corr{the \SD{} and \SDN{} sets}. To avoid favoring a specific task or model, the evaluation is conducted on the `MTL' baseline model and we select the set of weights from best $\dmtl{T}$ metrics.

\begin{table}[ht]%
    \centering
    \scriptsize
    \setlength{\tabcolsep}{0.003\textwidth}
    \captionsetup{font=footnotesize}
	\newcolumntype{H}{>{\setbox0=\hbox\bgroup}c<{\egroup}@{}}

    \subfloat[][\SD{} gridsearch]{
        \begin{tabular}{c c | c c | c }
			\toprule

			\multicolumn{2}{c|}{weights} & \hsem{}Semseg \uarr & \hdepth{}Depth \darr & \hdelta{}Delta \uarr \\

			$\omega_\text{S}$ & $\omega_\text{D}$ & \hsem{}\tiny{mIoU \%} & \hdepth{}\tiny{RMSE m} & \hdelta{}\tiny{$\dmtl{SD}$ \%} \\
			\midrule
			\tiny{1} & \tiny{1} & \perf{83.83}{0.15} & \perf{5.713}{0.060} & \perf{-0.35}{0.47} \\
			\tiny{1} & \tiny{10} & \perf{79.87}{0.21} & \uperf{5.708}{0.036} & \perf{-2.66}{0.40} \\
			\tiny{10} & \tiny{1} & \perf{86.20}{0.71} & \bperf{5.693}{0.055} & \perf{+1.30}{0.22} \\
			\oc{}\tiny{50} & \oc{}\tiny{1} & \oc{}\perf{87.73}{0.12} & \oc{}\perf{5.720}{0.029} & \oc{}\bperf{+1.89}{0.21} \\
            \tiny{100} & \tiny{1} & \perf{88.00}{0.20} & \perf{5.754}{0.030} & \uperf{+1.75}{0.17} \\
            \tiny{100} & \tiny{10} & \perf{86.13}{0.32} & \bperf{5.693}{0.039} & \perf{+1.18}{0.45} \\
            \tiny{200} & \tiny{1} & \uperf{88.13}{0.12} & \perf{5.790}{0.055} & \perf{+1.52}{0.45} \\
            \tiny{500} & \tiny{1} & \bperf{88.17}{0.15} & \perf{5.847}{0.043} & \perf{+1.04}{0.30} \\
			\bottomrule

		\end{tabular}
		\label{tab:SD_gs}}\\
    \subfloat[][\SDN{} gridsearch]{
        \begin{tabular}{c c c | c c H c | c }
            \toprule

            \multicolumn{3}{c|}{weights} & \hsem{}Semseg \uarr & \hdepth{}Depth \darr & \hdelta{}Delta \uarr & \hnormal{}Normals \darr & \hdelta{}Delta \uarr \\

            $\tiny{\omega_\text{S}}$ & $\omega_\text{D}$ & $\omega_\text{N}$ & \hsem{}\tiny{mIoU \%} & \hdepth{}\tiny{RMSE m} & \hdelta{}\tiny{$\dmtl{SD}$ \%} & \hnormal{}\tiny{mErr. °} & \hdelta{}\tiny{$\dmtl{SDN}$ \%} \\

            \midrule
            \tiny{1} & \tiny{1} & \tiny{1} & \perf{83.50}{0.20} & \perf{5.707}{0.058} & \perf{-0.50}{0.62} & \perf{23.03}{0.70} & \perf{-0.17}{0.64} \\
            \tiny{10} & \tiny{1} & \tiny{1} & \perf{86.53}{0.21} & \uperf{5.694}{0.032} & \perf{+1.40}{0.24} & \perf{22.98}{0.68} & \perf{+1.17}{0.93} \\
            \tiny{10} & \tiny{1} & \tiny{10} & \perf{86.63}{0.21} & \bperf{5.675}{0.050} & \perf{+1.63}{0.32} & \perf{22.61}{0.70} & \perf{+1.85}{0.80} \\
            \tiny{50} & \tiny{1} & \tiny{1} & \perf{87.73}{0.21} & \perf{5.706}{0.051} & \perf{+2.02}{0.32} & \perf{22.90}{0.71} & \perf{+1.69}{0.81} \\
            \tiny{50} & \tiny{1} & \tiny{10} & \perf{87.77}{0.15} & \perf{5.714}{0.065} & \perf{+1.96}{0.51} & \perf{22.56}{0.69} & \perf{+2.15}{0.86} \\
            \tiny{50} & \tiny{1} & \tiny{50} & \perf{87.73}{0.21} & \perf{5.701}{0.062} & \perf{+2.06}{0.42} & \perf{22.37}{0.70} & \perf{+2.49}{0.76} \\
            \tiny{100} & \tiny{1} & \tiny{1} & \perf{88.03}{0.15} & \perf{5.746}{0.030} & \perf{+1.84}{0.17} & \perf{22.95}{0.69} & \perf{+1.49}{0.92} \\
            \tiny{100} & \tiny{1} & \tiny{10} & \perf{87.97}{0.15} & \perf{5.714}{0.048} & \perf{+2.08}{0.33} & \perf{22.59}{0.69} & \perf{+2.19}{0.79} \\
            \tiny{100} & \tiny{1} & \tiny{50} & \perf{88.00}{0.20} & \perf{5.717}{0.048} & \perf{+2.07}{0.34} & \perf{22.40}{0.71} & \perf{+2.45}{0.99} \\
            \oc{}\tiny{100} & \oc{}\tiny{1} & \oc{}\tiny{100} & \oc{}\perf{88.07}{0.15} & \oc{}\perf{5.696}{0.038} & \oc{}\perf{+2.30}{0.30} & \oc{}\bperf{22.29}{0.70} & \oc{}\bperf{+2.75}{1.04} \\
            \tiny{150} & \tiny{1} & \tiny{10} & \uperf{88.10}{0.20} & \perf{5.752}{0.059} & \perf{+1.83}{0.41} & \perf{22.59}{0.70} & \perf{+2.01}{0.86} \\
            \tiny{150} & \tiny{1} & \tiny{50} & \uperf{88.10}{0.20} & \perf{5.738}{0.039} & \perf{+1.95}{0.26} & \perf{22.41}{0.70} & \perf{+2.35}{0.99} \\
            \tiny{150} & \tiny{1} & \tiny{100} & \bperf{88.13}{0.15} & \perf{5.732}{0.037} & \perf{+2.02}{0.24} & \uperf{22.31}{0.71} & \uperf{+2.54}{0.94} \\
            \bottomrule

        \end{tabular}%
        \label{tab:SDN_gs}}

    \caption{Performance of the `MTL' baseline model (cf.~\cref{fig:archi}) for different sets of multi-task weights on \vk. There are important remarks. First, uniform weighting is far from optimal. Second, best $\dmtl{T}$ does not always equate to optimal individual metrics as shown by the results in \textbf{bold}.
	Ultimately, to avoid favoring a single task, we use the set of weights with highest $\dmtl{T}$ metric for all models, as highlighted in \colorbox{gray!10}{gray}.}
	\label{tab:mtl-gridsearch}

\end{table}

\subsection{Main results}
\label{supp:main_results}
\subsubsection{General architectures.}%
\begin{figure*}[ht!]
	\centering
	\scriptsize
	\captionsetup{font=footnotesize}
	\resizebox{\linewidth}{!}{%
    	\begin{tabular}{cc|cc|cc}

    		\multicolumn{2}{c}{\includegraphics[width=0.3\linewidth,valign=m]{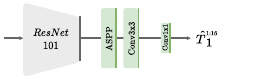}}
    		& \multicolumn{2}{c}{\includegraphics[width=0.3\linewidth,valign=m]{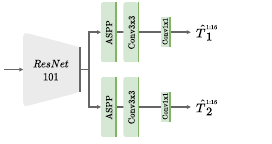}}
    		& \multicolumn{2}{c}{\includegraphics[width=0.4\linewidth,valign=m]{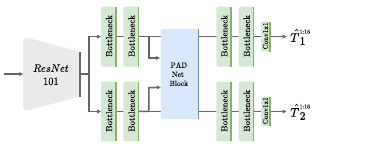}} \\
    		\multicolumn{2}{c}{STL \cite{Vandenhende2021}} & \multicolumn{2}{c}{MTL \cite{Vandenhende2021}} & \multicolumn{2}{c}{PAD-Net \cite{xu2018padnet}} \\ [1ex]

    		\multicolumn{3}{c}{\includegraphics[width=0.5\linewidth,valign=m]{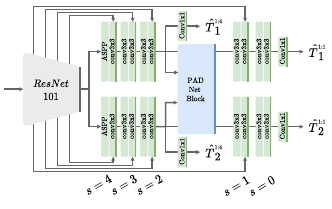}}
    		& \multicolumn{3}{c}{\includegraphics[width=0.5\linewidth,valign=m]{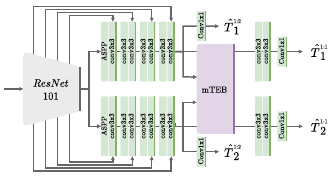}} \\
    		\multicolumn{3}{c}{\threewaysPADNet \cite{hoyer2021ways}} & \multicolumn{3}{c}{Ours} \\

    	\end{tabular}
    }
	\caption{\textbf{General architectures.} For clarity, we only visualize two tasks in the multi-task networks. While the encoder is identical, models differ in their decoder architecture, with PAD-Net, \threewaysPADNet{} and Ours using dedicated tasks exchange blocks.
	}
	\label{fig:archi}
\end{figure*}

In~\cref{fig:archi} we show the general architectures (\ie, considering depth supervision), including the STL, MTL and PAD-Net models~\cite{hoyer2021ways}, \threewaysPADNet{}~\cite{hoyer2021ways} and Ours.
Based on the same encoder taken from a pretrained ResNet-101 backbone~\cite{he2016deep}, those multi-task networks differ only in decoder design.
We observe improvements in all tasks using both Atrous Spatial Pyramid Pooling (ASPP) and UNet-like connections as done in~\cite{hoyer2021ways} (cf. \threewaysPADNet{} \textit{vs.} PAD-Net). %

\subsubsection{SDE architectures.} %
\begin{figure}[ht!]
	\centering
	\scriptsize
	\captionsetup{font=footnotesize}
    \includegraphics[width=0.9\linewidth,valign=m]{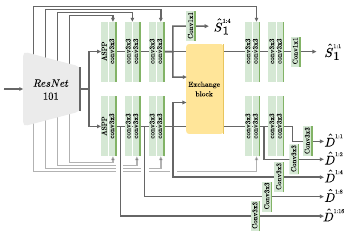}
	\caption{\textbf{Architecture for Self\corr{-supervised} Depth Estimation (SDE).} To accommodate monocular depth on `Cityscapes SDE', we follow the setup of~\cite{hoyer2021ways} with added intermediate depth supervision ($\hat{D}^{1:\textbf{x}}$). For the two variants in the SDE setup, we use the above architecture, replacing the `exchange block' with the \corr{desired} one.}
	\label{fig:archi_SDE}
\end{figure}

To allow monocular depth estimation in `MTL for segmentation', %
we adopt the setup of \cite{hoyer2021ways} where intermediate depth estimation from pair of consecutive frames is supervised by a photometric reconstruction loss~\cite{godard2019digging}.
\cref{fig:archi_SDE} shows the architecture used for all 3-ways variants for the semantics training with SDE. %
Variants consist of swapping the yellow `Exchange block' with either `PAD-Net block' (\threewaysPADNet{}) or our `mTEB' (\threewaysOurs{}).

To train, we use $1.0e{-5}$, $5.0e{-5}$, and $1.0e{-6}$ as learning rates for the encoder, decoder and pose estimation network respectively.
The training strategy is similar to our other MTL setups, only this time we initialize all models with weights from a single-branch model trained on self-supervised depth estimation (cf. \cite{hoyer2021ways}).
Since the depth loss differs from the supervised ones, %
we do \textit{not} apply the weighting found for \SD{} but instead resort to uniform weighting for direct comparison to \cite{hoyer2021ways}.

\subsection{MTL for Unsupervised Domain Adaptation}
\label{supp:mtl_uda}

\newcommand{\mc}[1]{\ensuremath{\mathcal{#1}}}
\newcommand{\m}[1]{\ensuremath{\bm{#1}}}
\def\mI{\mc{I}}
\def\mQ{\mc{Q}}
\def\mP{\mc{P}}
\def\mE{\mc{E}}
\def\domain{\mc{X}}
\def\loss{\mc{L}}
\def\vx{\m{x}}

\subsubsection{Architecture and training.}
\begin{figure}[!h]
    \captionsetup{font=footnotesize}
	\centering
	\includegraphics[width=0.9\linewidth]{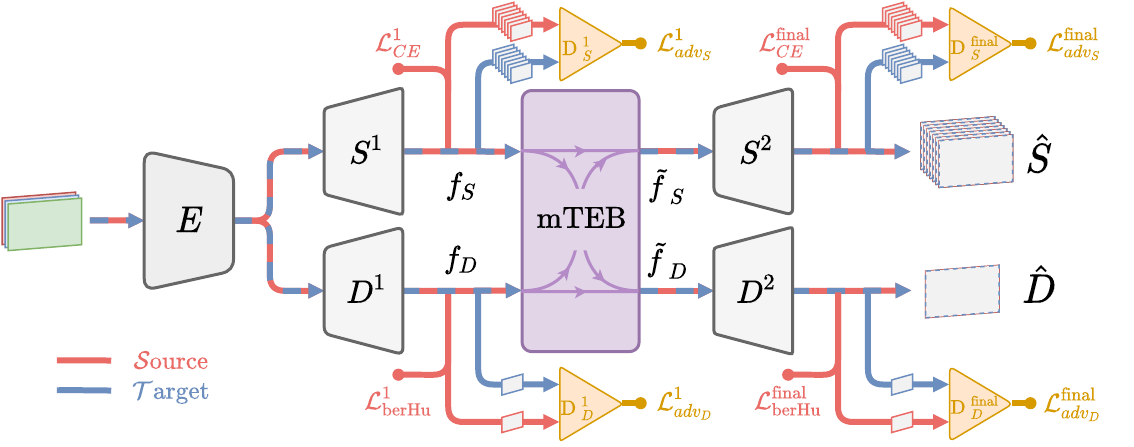}
	\caption{\textbf{Multi-task UDA.}
	Arrows indicating data flows are drawn in either red (source), blue (target) or a mix (both).
	Additional discriminators (shown as yellow triangles) are jointly trained with our multi-task model.
	}
	\label{fig:mtlda}
\end{figure}

\cref{fig:mtlda} illustrates our adversarial learning scheme with source/target data flows for multi-task UDA.
We consider the two-task `S-D' setup.
As explained in our paper, %
domain alignment is made possible with output-level DA adversarial training.
In our work, alignment is done at both intermediate and final output-levels.

We follow the strategies introduced in~\cite{vu2019advent,wang2021semisupervised}. Discriminators D are train on the source dataset $\domain_{src}$ and target dataset $\domain_{trg}$ by minimizing the binary classification loss:
\begin{multline}
    \loss_{\text{D}} = \frac{1}{|\domain_{src}|}\sum_{\vx_{src}\in\domain_{src}}{\loss_\text{BCE}(\text{D}(\mQ_{\vx_{src}}),1)} +\\ \frac{1}{|\domain_{trg}|}\sum_{\vx_{trg}\in\domain_{trg}} \loss_\text{BCE}(\text{D}(\mQ_{\vx_{trg}}),0),
\end{multline}
where $\loss_\text{BCE}$ is the Binary Cross-Entropy loss, and $\mQ_{\vx}$ stands for either segmentation output $\mQ_{\vx}^S$ or depth output $\mQ_{\vx}^D$ of the network.
To compete with the discriminators, the main MTL network is additionally trained with the adversarial losses $\loss_{adv}$, written as:
\begin{equation}
    \loss_{adv} = \frac{1}{|\domain_{trg}|}\sum_{\vx_{trg}\in\domain_{trg}} \loss_\text{BCE}(\text{D}(\mQ_{\vx_{trg}}),1).
\end{equation}
The final MTL-UDA loss becomes:
\begin{multline}
	\mathcal{L}_{\text{MTL-UDA}} = \frac{1}{\lvert S \rvert}\sum_{s \in S}{\sum_{t \in T}({\omega_{t}\mathcal{L}^s_t + \lambda_{adv}\mathcal{L}_{adv_{t}}^\text{s}}})\, \\ + {\sum_{t \in T}({\omega_{t}\mathcal{L}^\text{final}_t} + \lambda_{adv}\mathcal{L}_{adv_{t}}^\text{final}}),
\end{multline}
where $\lambda_{adv}$ is used to weight the adversarial losses and is set to $5.0e{-3}$.

For \textbf{segmentation} alignment, we use ``weighted self-information'' map~\cite{vu2019advent} computed from the softmax segmentation output $\mP_{\vx}$ with the formula:
\begin{equation}
    \mQ_{\vx}^{S} = -\mP_{\vx} \odot \log(\mP_{\vx}).
\end{equation}

For \textbf{depth} alignment, we normalize the depth-map outputs using the source's min and max depth values, and directly align the continuous normalized maps $\mQ_{\vx}^D$~\cite{wang2021semisupervised}.

\subsubsection{Class mapping.}

To allow compatible semantics in the \vk$\mapsto$\allowbreak\cs{} setup, we adopt the mapping of~\cref{tab:vktocs_mapping}.

\begin{table}
	\centering
	\scriptsize
	\captionsetup{font=footnotesize}
	\begin{tabular}{l l | l l}
		\vk         & mapped            & \cs           & mapped \\
		\toprule
		terrain     & \textit{ignore}   & road          & road \\
		sky         & sky               & sidewalk      & \textit{ignore} \\
		tree        & vegetation        & building      & building \\
		vegetation  & vegetation        & wall          & vegetation \\
		building    & building          & fence         & \textit{ignore} \\
		road        & road              & pole          & pole \\
		guardrail   & \textit{ignore}   & light         & light \\
		sign        & sign              & sign          & sign \\
		light       & light             & vegetation    & vegetation \\
		pole        & pole              & sky           & sky \\
		misc        & \textit{ignore}   & person        & \textit{ignore} \\
		truck       & vehicle           & rider         & \textit{ignore} \\
		car         & vehicle           & car           & vehicle \\
		van         & vehicle           & bus           & vehicle \\
		            &                   & mbike         & \textit{ignore} \\
		            &                   & bike          & \textit{ignore} \\
	\end{tabular}
	\caption{Class mapping for \vk$\mapsto$\cs{} DA setup.}
	\label{tab:vktocs_mapping}
\end{table}

\section{Additional results}
\label{sec:supp-results}
\sloppy
\cref{fig:mtl_quali_sy,fig:mtl_quali_vk,fig:mtl_quali_cs} show additional qualitative results for \sy{}, \vk{}, and \cs{}, respectively. Comparing \textit{Ours} with \textit{PAD-Net} show an evident segmentation improvement on thin elements such as poles or pedestrians in \cref{fig:mtl_quali_sy,fig:mtl_quali_cs} with significantly sharper results for depth and normals across setups.
Comparing against \textit{\threewaysPADNet{}} is harder due to their high scores (cf. main paper Tab. {\color{red}1}).

\newcommand{\img}[1]{\includegraphics[width=0.155\linewidth,valign=m]{#1}}

\begin{figure*}
    \scriptsize
    \newcommand{\vizinput}[1]{ %
        \img{viz/other/synthia_#1_rgb_gt.png} & %
    	& \img{viz/other/synthia_#1_semseg_gt.png} %
    	& \img{viz/other/synthia_#1_depth_gt.png} %
    	& \img{viz/other/synthia_#1_semseg_gt.png} %
    	& \img{viz/other/synthia_#1_depth_gt.png} %
    	& \img{viz/other/synthia_#1_normals_gt.png} %
    }
    \newcommand{\vizmethodSD}[2]{ %
        \img{viz/other/synthia_resnet101_#2_SD_#1_semseg_0.png} & %
    	\img{viz/other/synthia_resnet101_#2_SD_#1_depth_0.png} %
    }
    \newcommand{\vizmethodSDN}[2]{ %
    	\img{viz/other/synthia_resnet101_#2_SDN_#1_semseg_0.png} &%
    	\img{viz/other/synthia_resnet101_#2_SDN_#1_depth_0.png} &%
    	\img{viz/other/synthia_resnet101_#2_SDN_#1_normals_0.png} %
    }
    \newcommand{\vizmethodSDSDN}[2]{ %
        \vizmethodSD{#1}{#2}&\vizmethodSDN{#1}{#2}
    }
    \resizebox{\linewidth}{!}{%
    \begin{tabular}{rr cc?ccc}
    	& & \multicolumn{2}{c?}{\SD{}} & \multicolumn{3}{c}{\SDN{}} \\[1ex]

    	&&\hsem{}\textbf{Segmentation}&\hdepth{}\textbf{Depth}&\hsem{}\textbf{Segmentation}&\hdepth{}\textbf{Depth}&\hnormal{}\textbf{Normal}\\

    	\vizinput{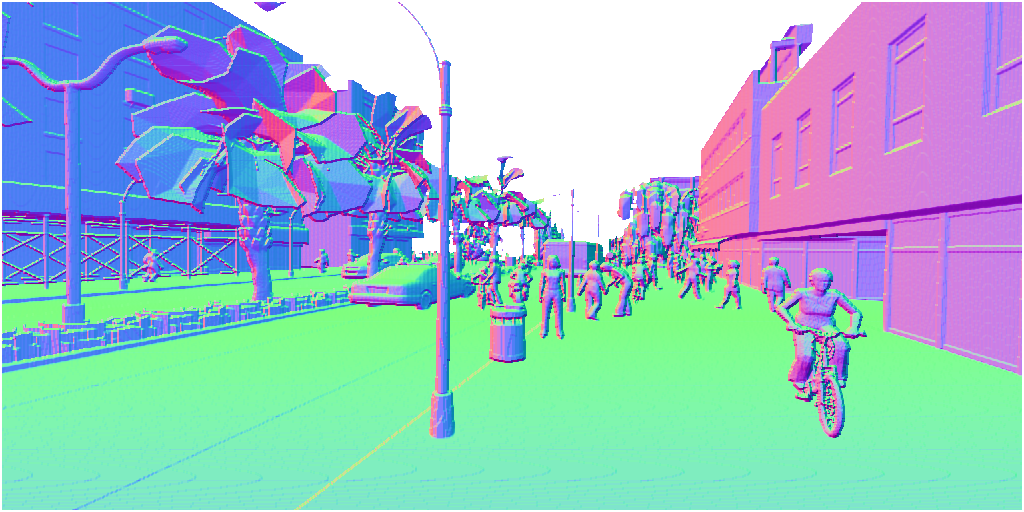} \\[1ex]
    	PAD-Net~\cite{xu2018padnet}&&\vizmethodSDSDN{RAND_CITYSCAPES_RGB_0001311}{padnet}\\
    	\threewaysPADNet{}~\cite{hoyer2021ways}&&\vizmethodSDSDN{RAND_CITYSCAPES_RGB_0001311}{3ways}\\
    	Ours&&\vizmethodSDSDN{RAND_CITYSCAPES_RGB_0001311}{ours+s1}\\[3em]

    	\vizinput{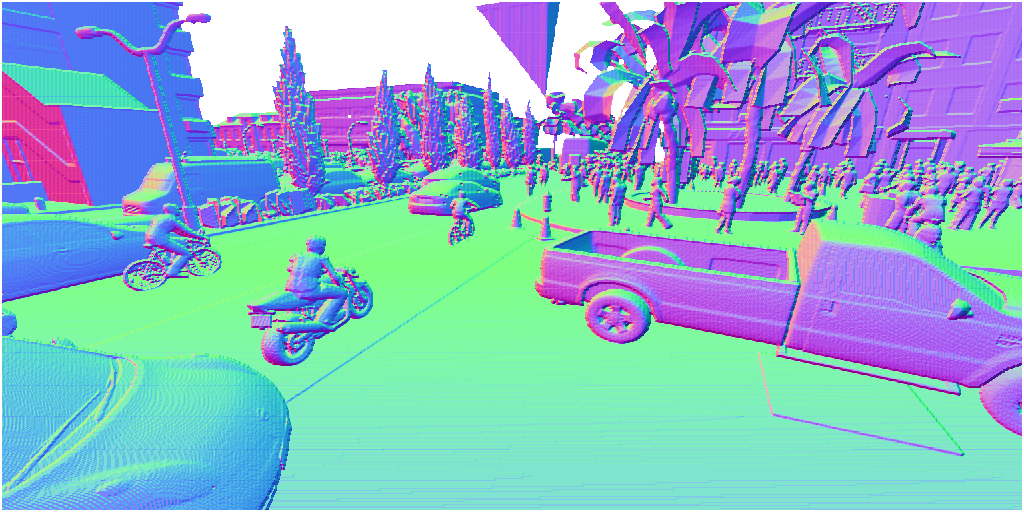} \\[1ex]
    	PAD-Net~\cite{xu2018padnet}&&\vizmethodSDSDN{RAND_CITYSCAPES_RGB_0005330}{padnet}\\
    	\threewaysPADNet{}~\cite{hoyer2021ways}&&\vizmethodSDSDN{RAND_CITYSCAPES_RGB_0005330}{3ways}\\
    	Ours&&\vizmethodSDSDN{RAND_CITYSCAPES_RGB_0005330}{ours+s1}\\

    \end{tabular}}
    \caption{\textbf{Qualitative results on \sy{}.} Overall, \textit{Ours} produces better and sharper. Comparing visually against \threewaysPADNet{} is harder due to high scores.}
    \label{fig:mtl_quali_sy}
\end{figure*}

\begin{figure*}[ht!]
    \scriptsize
    \resizebox{\linewidth}{!}{%
    \begin{tabular}{rr cc?ccc}
    	& & \multicolumn{2}{c?}{\SD{}} & \multicolumn{3}{c}{\SDN{}} \\[1ex]

    	&&\hsem{}\textbf{Segmentation}&\hdepth{}\textbf{Depth}&\hsem{}\textbf{Segmentation}&\hdepth{}\textbf{Depth}&\hnormal{}\textbf{Normal}\\

    	  \img{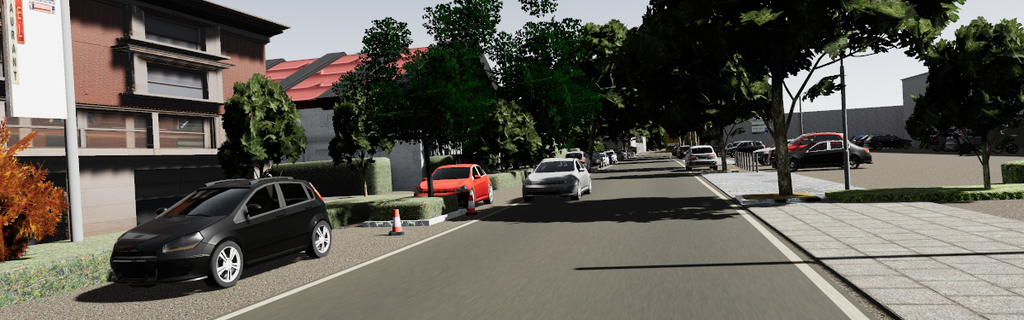} &
    	& \img{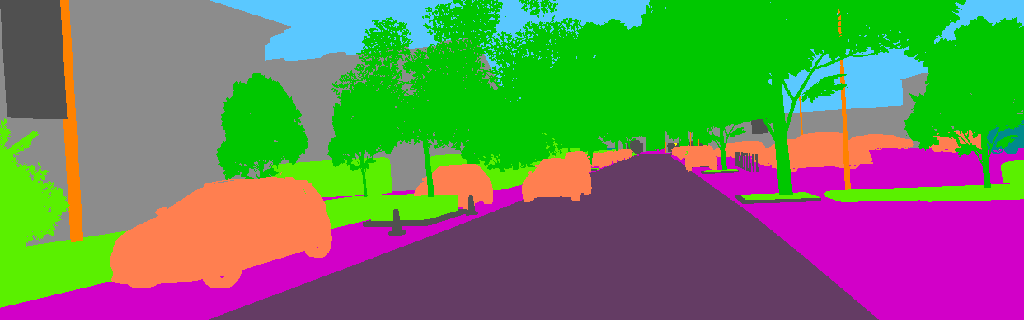}
    	& \img{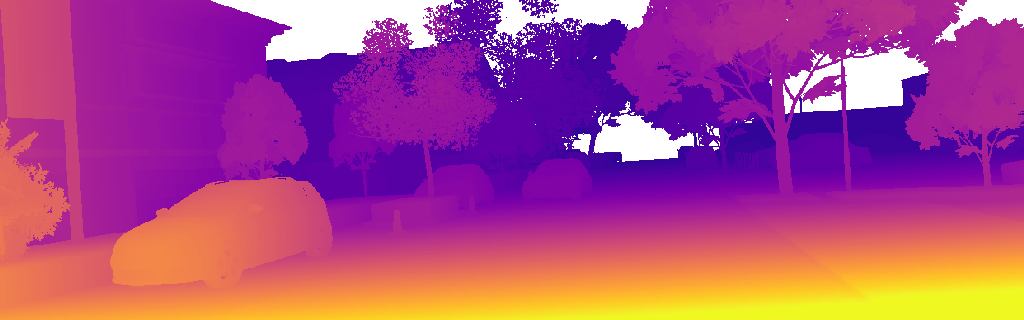}
    	& \img{viz/other/vkitti2_Scene01_15-deg-left_frames_rgb_Camera_0_rgb_00053_semseg_gt.png}
    	& \img{viz/other/vkitti2_Scene01_15-deg-left_frames_rgb_Camera_0_rgb_00053_depth_gt.png}
    	& \img{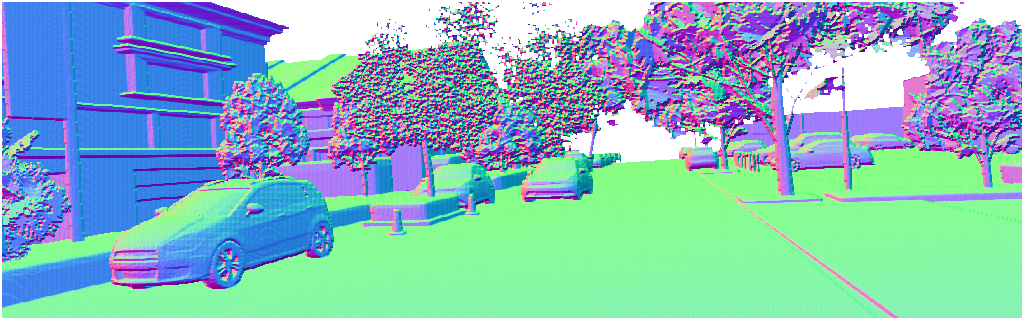} \\[1ex]

    	PAD-Net~\cite{xu2018padnet}&
    	& \img{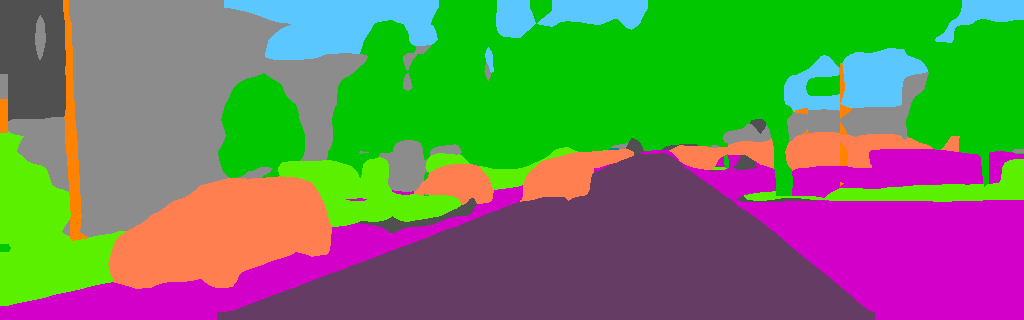}
    	& \img{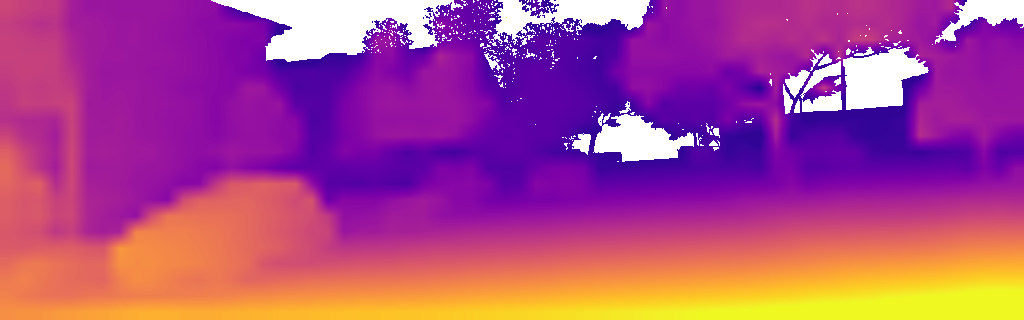}
    	& \img{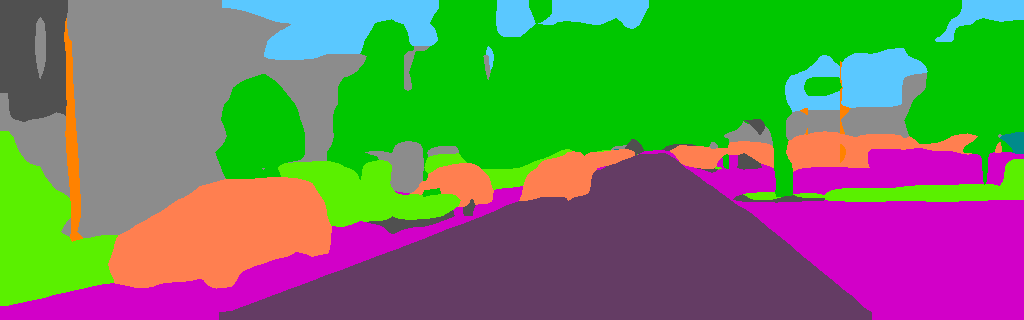}
    	& \img{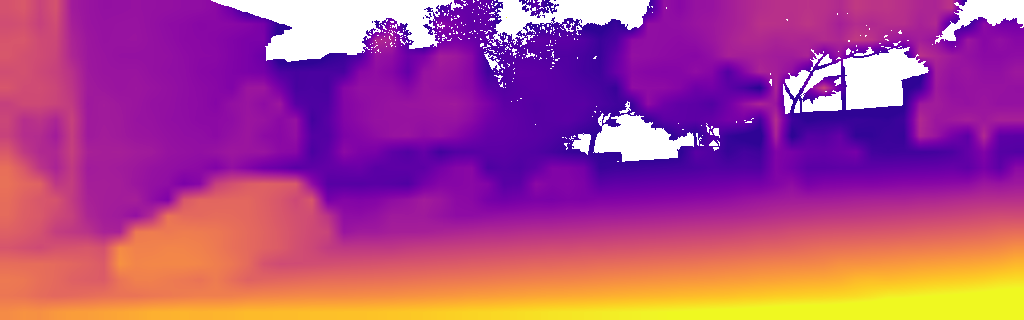}
    	& \img{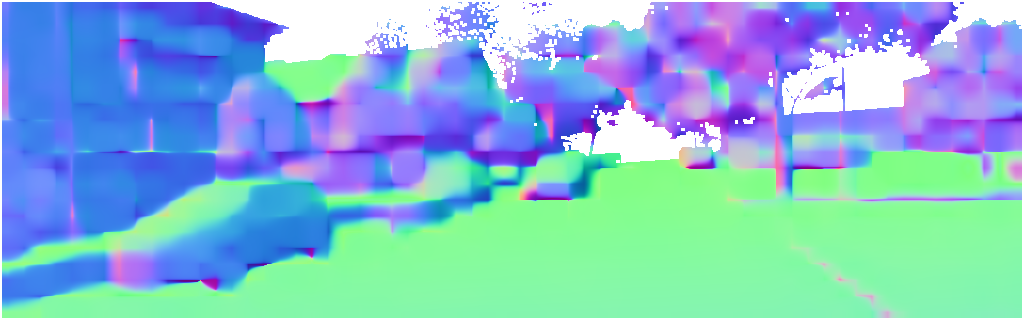} \\

    	\threewaysPADNet{}~\cite{hoyer2021ways} &
    	& \img{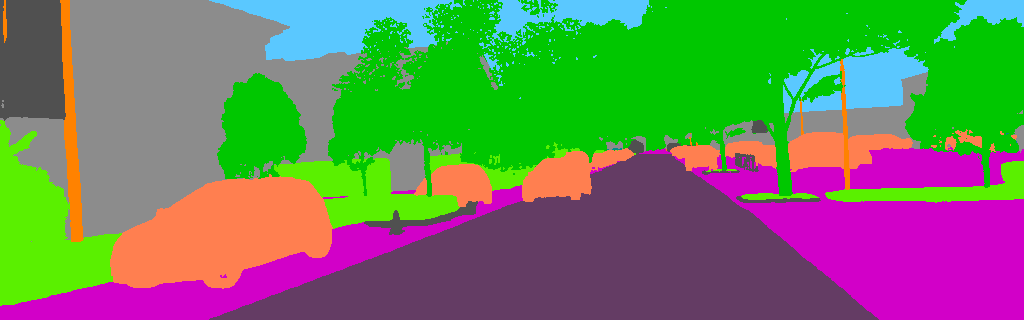}
    	& \img{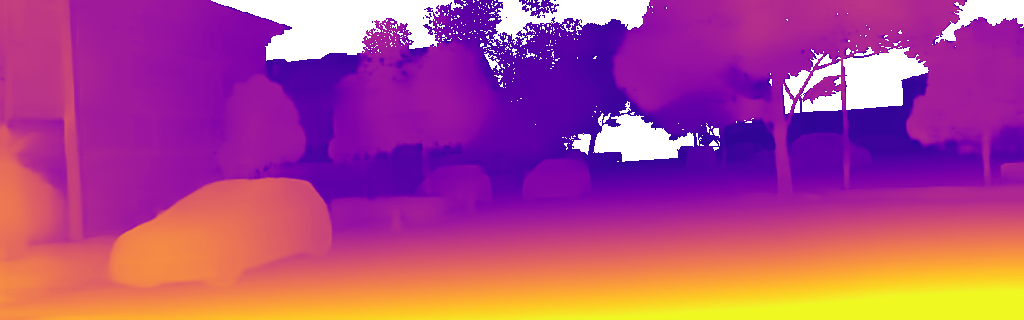}
    	& \img{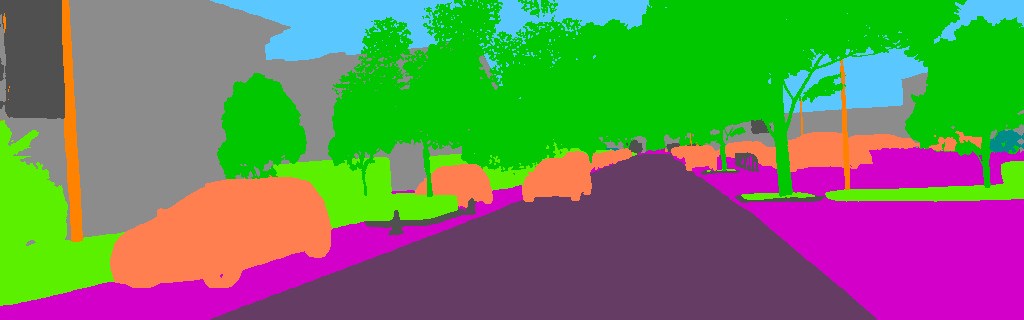}
    	& \img{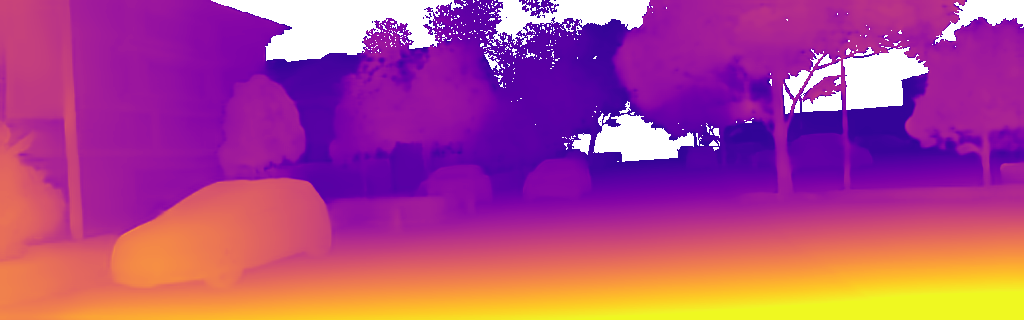}
    	& \img{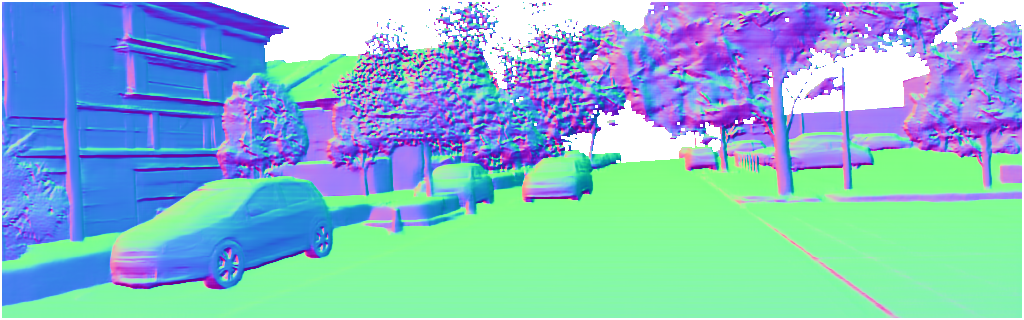} \\

    	Ours &
    	& \img{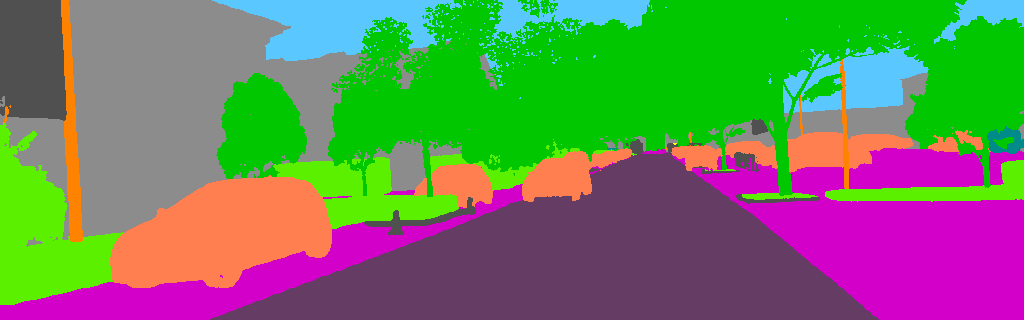}
    	& \img{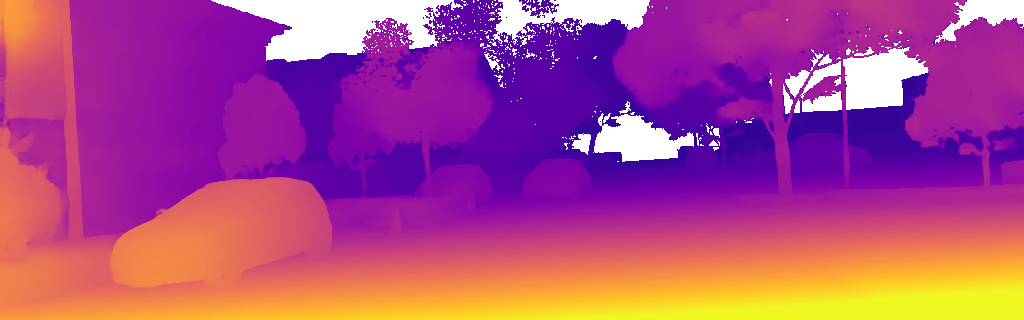}
    	& \img{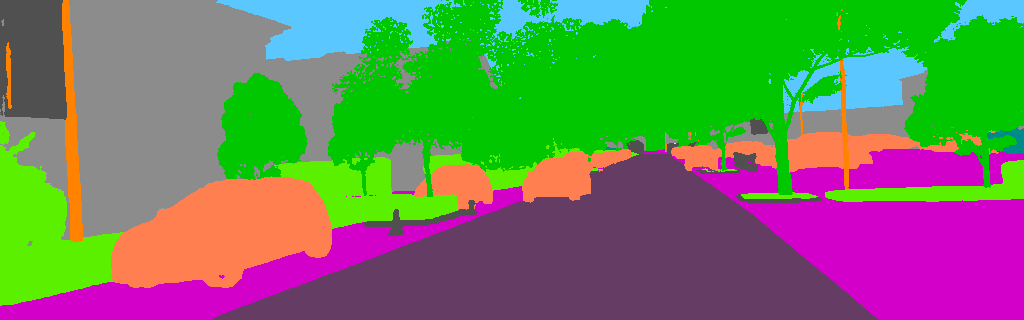}
    	& \img{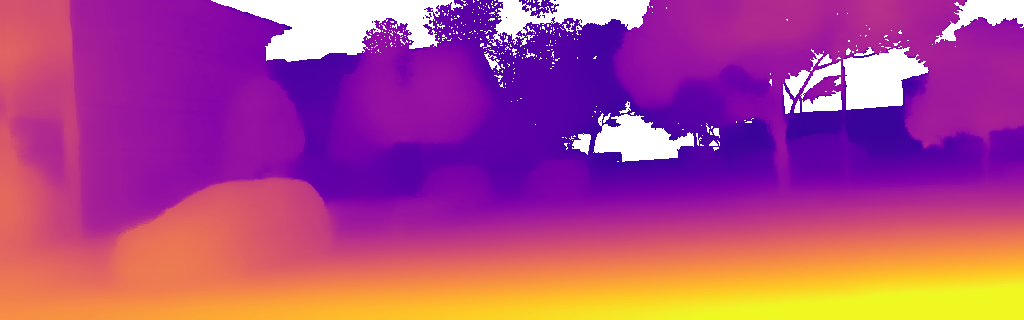}
    	& \img{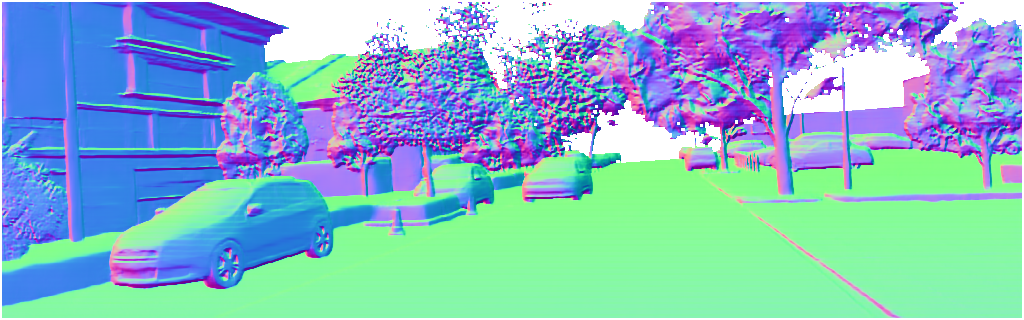} \\[3em]

    	  \img{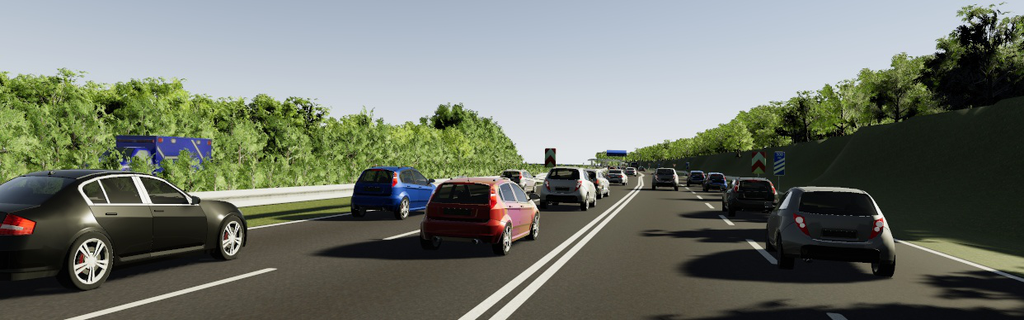} &
    	& \img{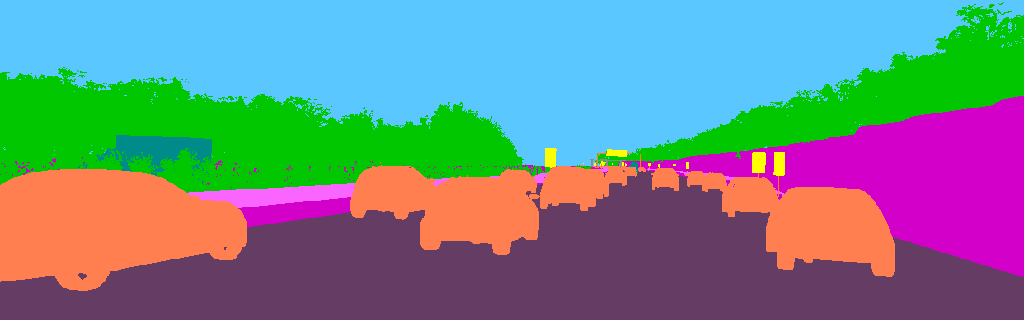}
    	& \img{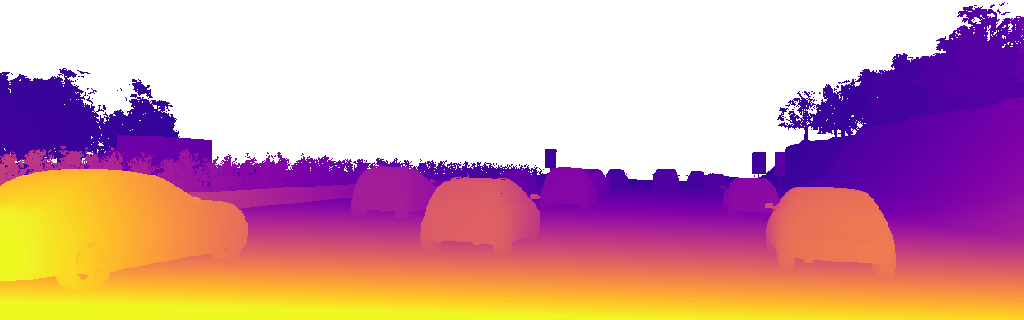}
    	& \img{viz/other/vkitti2_Scene20_15-deg-left_frames_rgb_Camera_0_rgb_00017_semseg_gt.png}
    	& \img{viz/other/vkitti2_Scene20_15-deg-left_frames_rgb_Camera_0_rgb_00017_depth_gt.png}
    	& \img{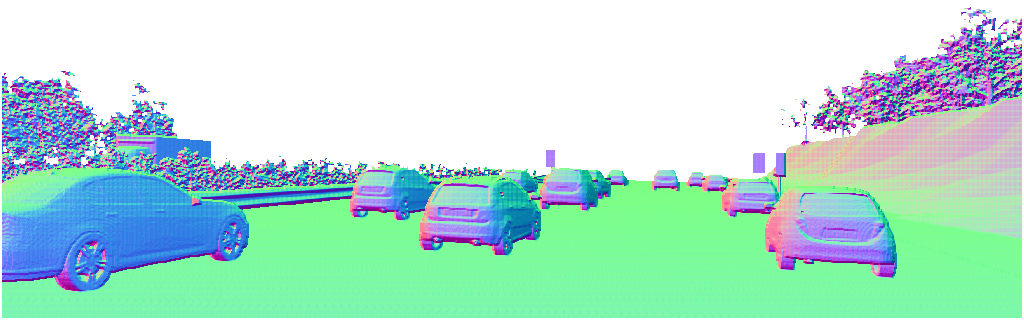} \\[1ex]

    	PAD-Net~\cite{xu2018padnet}&
    	& \img{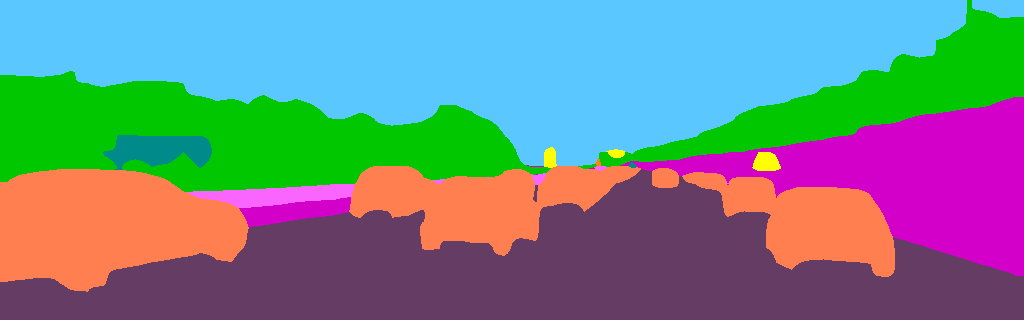}
    	& \img{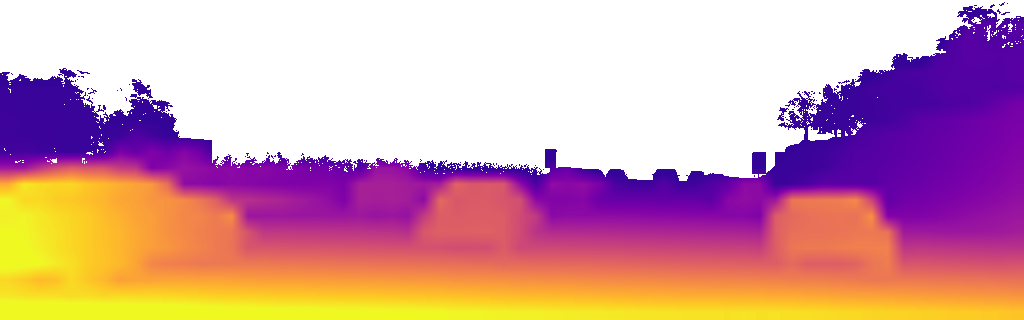}
    	& \img{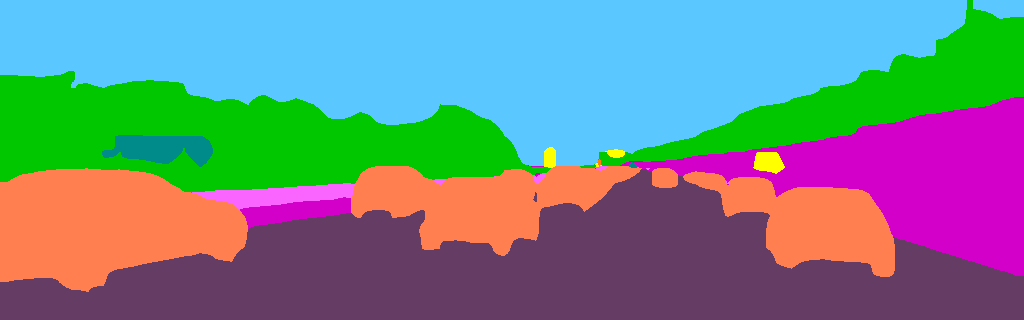}
    	& \img{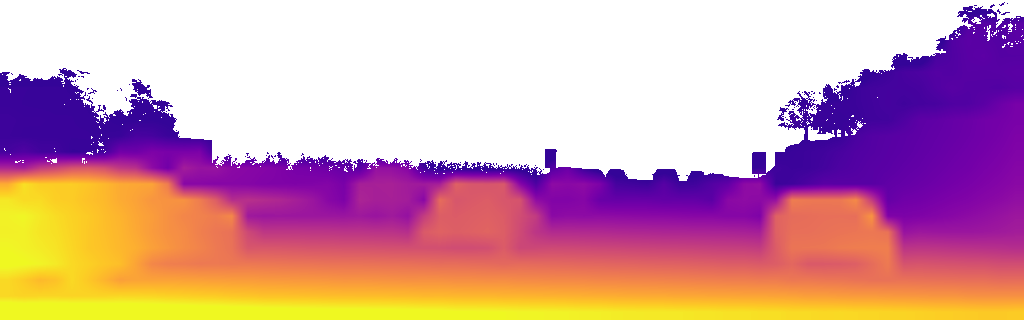}
    	& \img{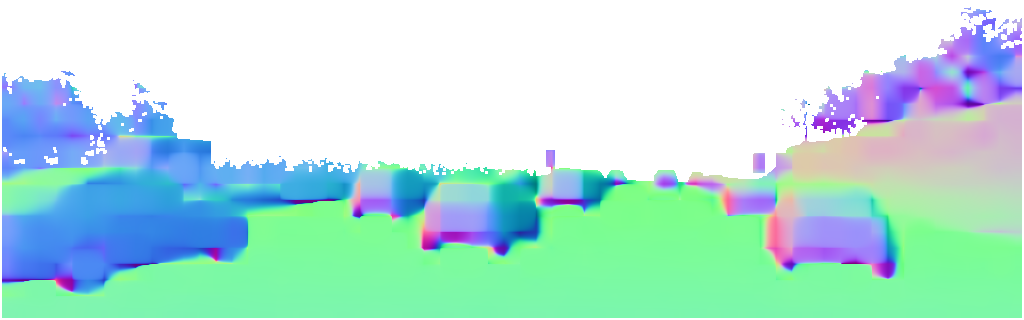} \\

    	\threewaysPADNet{}~\cite{hoyer2021ways} &
    	& \img{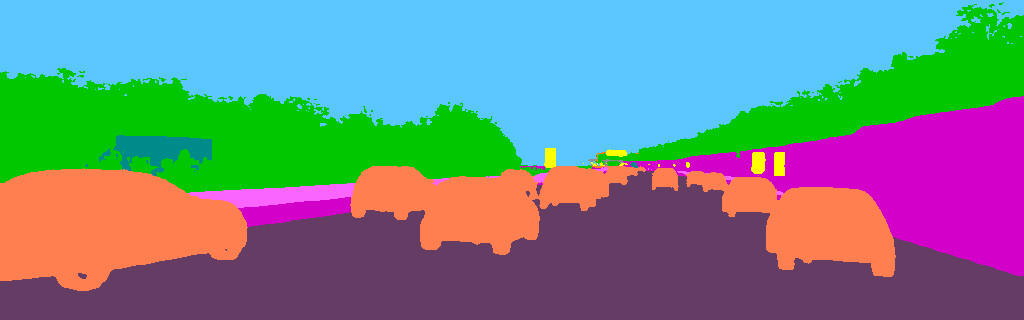}
    	& \img{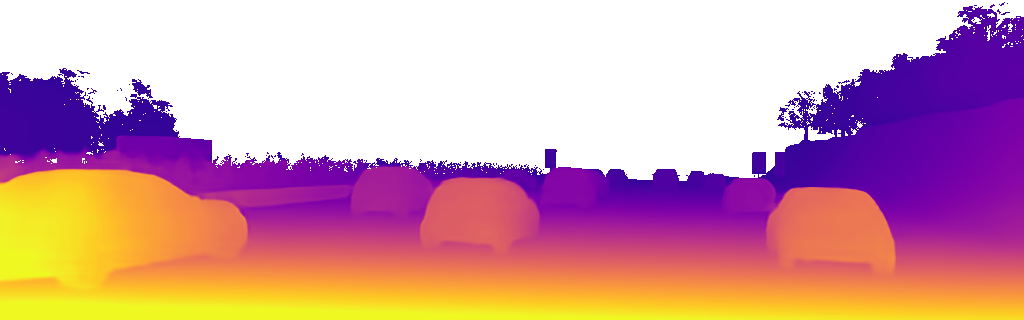}
    	& \img{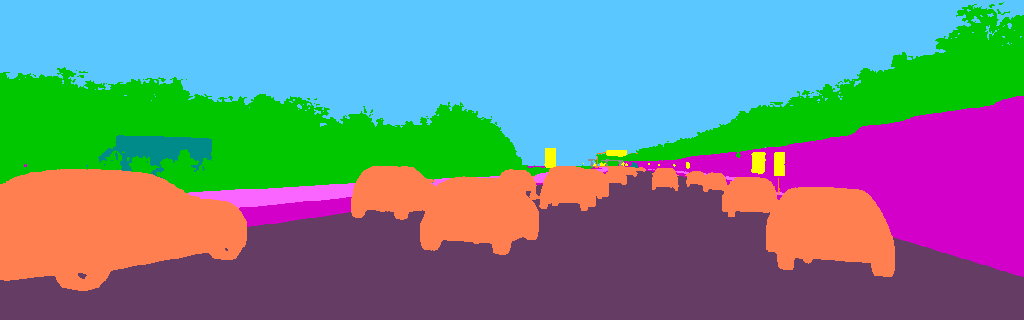}
    	& \img{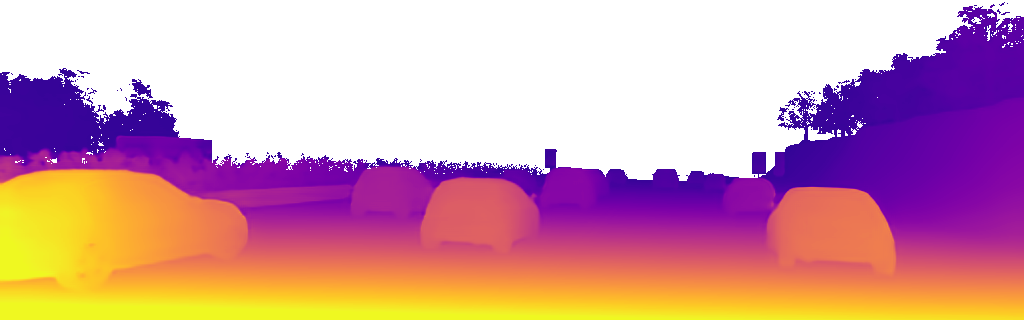}
    	& \img{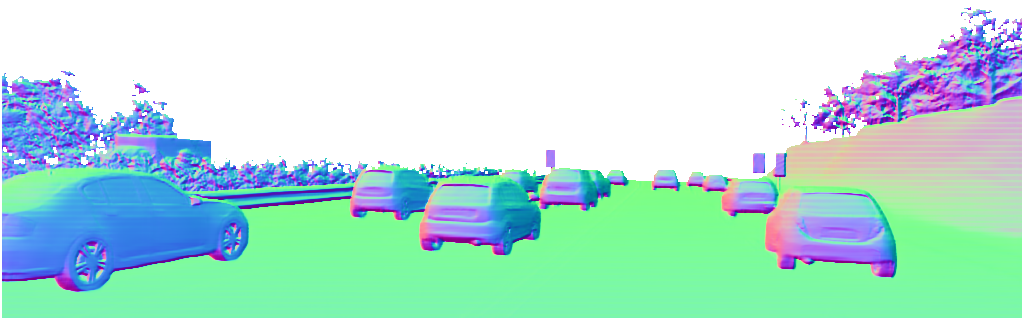} \\

    	Ours &
    	& \img{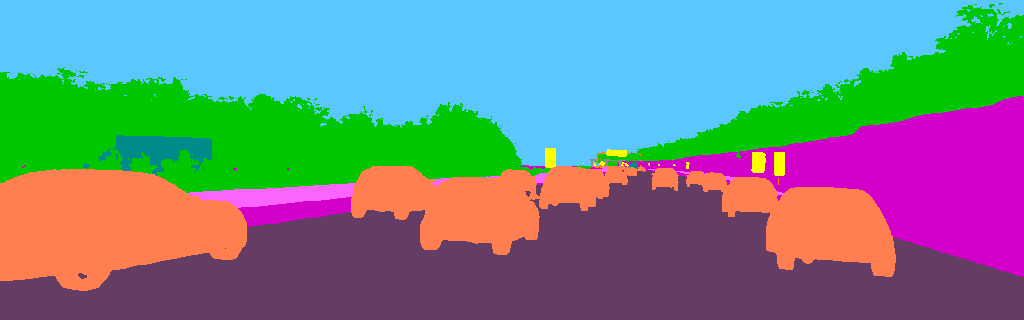}
    	& \img{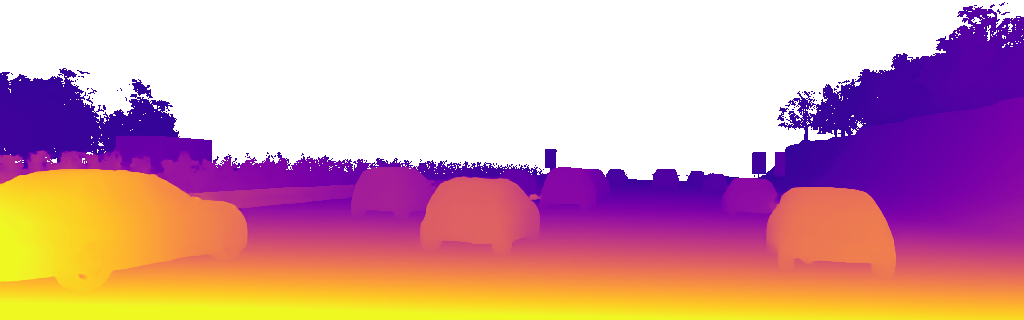}
    	& \img{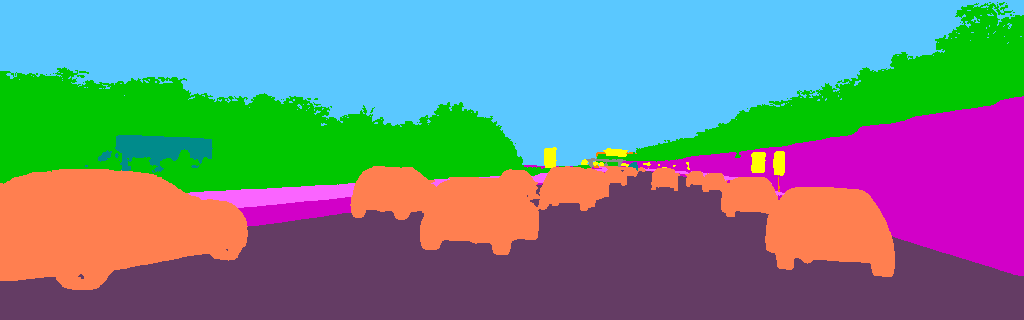}
    	& \img{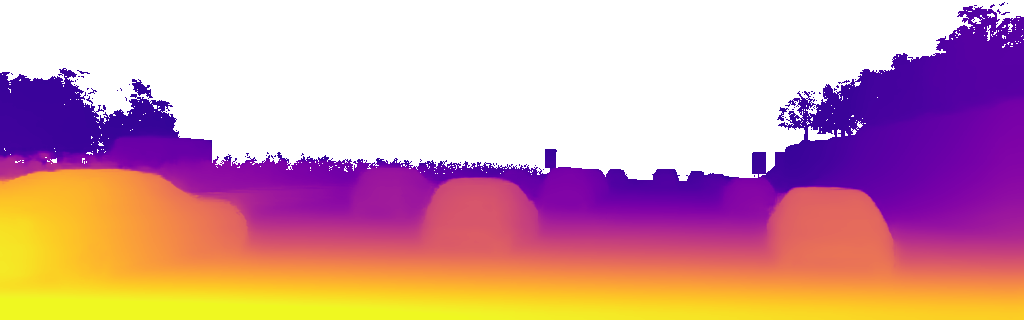}
    	& \img{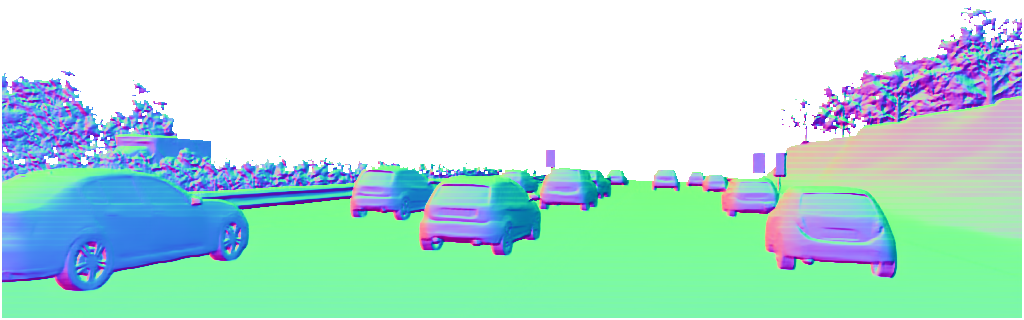} \\

    \end{tabular}}
    \caption{\textbf{Qualitative results on \vk{}}. Overall, \textit{Ours} produce\corr{s} better and sharper. Comparing against \threewaysPADNet{} is harder visually due to high scores.} %
    \label{fig:mtl_quali_vk}
\end{figure*}

\begin{figure*}
    \scriptsize
    \newcommand{\vizinput}[1]{ %
        \img{viz/other/cityscapes_leftImg8bit_#1_rgb_gt.png} & %
    	& \img{viz/other/cityscapes_leftImg8bit_#1_semseg_gt.png} %
    	& \img{viz/other/cityscapes_leftImg8bit_#1_depth_gt.png} %
    	& \img{viz/other/cityscapes_leftImg8bit_#1_semseg_gt.png} %
    	& \img{viz/other/cityscapes_leftImg8bit_#1_depth_gt.png} %
    	& \img{viz/other/cityscapes_leftImg8bit_#1_normals_gt.png} %
    }
    \newcommand{\vizmethodSD}[2]{ %
        \img{viz/other/cityscapes_resnet101_#2_SD_leftImg8bit_#1_semseg_0.png} & %
    	\img{viz/other/cityscapes_resnet101_#2_SD_leftImg8bit_#1_depth_0.png} %
    }
    \newcommand{\vizmethodSDN}[2]{ %
    	\img{viz/other/cityscapes_resnet101_#2_SDN_leftImg8bit_#1_semseg_0.png} &%
    	\img{viz/other/cityscapes_resnet101_#2_SDN_leftImg8bit_#1_depth_0.png} &%
    	\img{viz/other/cityscapes_resnet101_#2_SDN_leftImg8bit_#1_normals_0.png} %
    }
    \newcommand{\vizmethodSDSDN}[2]{ %
        \vizmethodSD{#1}{#2}&\vizmethodSDN{#1}{#2}
    }

    \resizebox{\linewidth}{!}{%
    \begin{tabular}{rr cc?ccc}
    	& & \multicolumn{2}{c?}{\SD{}} & \multicolumn{3}{c}{\SDN{}} \\[1ex]

    	&&\hsem{}\textbf{Segmentation}&\hdepth{}\textbf{Depth}&\hsem{}\textbf{Segmentation}&\hdepth{}\textbf{Depth}&\hnormal{}\textbf{Normals}\\

    	\vizinput{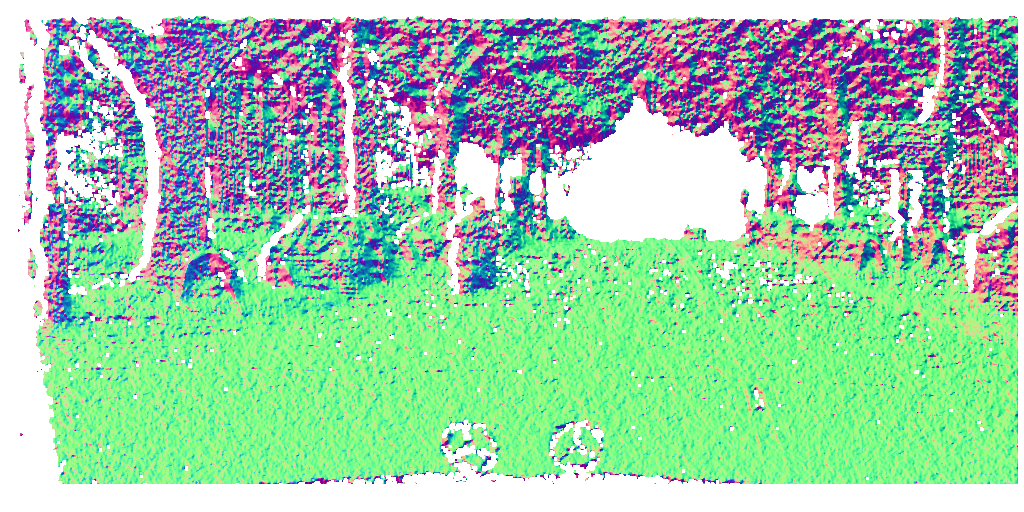} \\[1ex]
    	PAD-Net~\cite{xu2018padnet}&&\vizmethodSDSDN{small_val_munster_munster_000094_000019_leftImg8bit}{padnet}\\
    	\threewaysPADNet{}~\cite{hoyer2021ways}&&\vizmethodSDSDN{small_val_munster_munster_000094_000019_leftImg8bit}{3ways}\\
    	Ours&&\vizmethodSDSDN{small_val_munster_munster_000094_000019_leftImg8bit}{ours+s1}\\[3em]

    	\vizinput{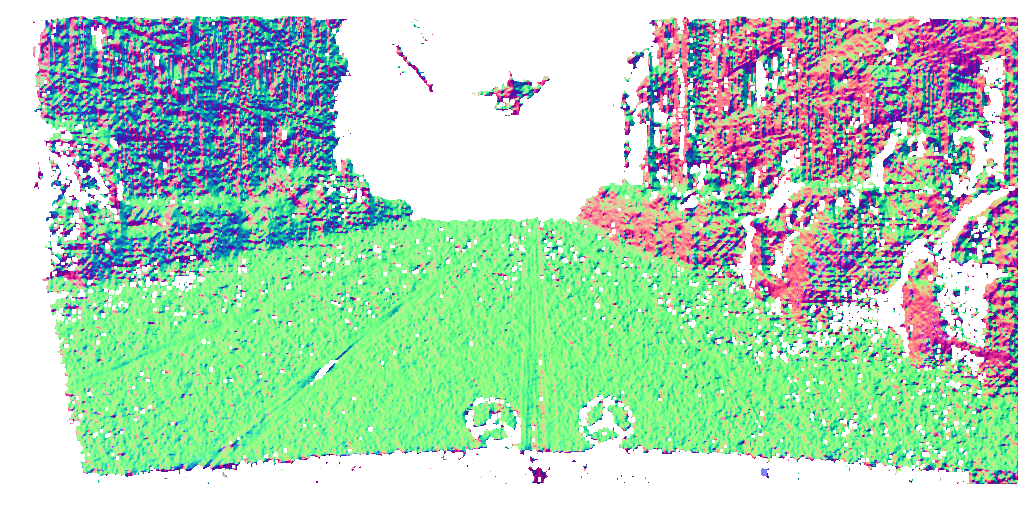} \\[1ex]
    	PAD-Net~\cite{xu2018padnet}&&\vizmethodSDSDN{small_val_frankfurt_frankfurt_000001_033655_leftImg8bit}{padnet}\\
    	\threewaysPADNet{}~\cite{hoyer2021ways}&&\vizmethodSDSDN{small_val_frankfurt_frankfurt_000001_033655_leftImg8bit}{3ways}\\
    	Ours&&\vizmethodSDSDN{small_val_frankfurt_frankfurt_000001_033655_leftImg8bit}{ours+s1}\\
    \end{tabular}}
    \caption{\textbf{Qualitative results on \cs{}}. Overall, \textit{Ours} produce\corr{s} better and sharper. Comparing visually against \threewaysPADNet{} is harder due to high scores.}
    \label{fig:mtl_quali_cs}
\end{figure*}

\newpage
\twocolumn[
\begin{center}
\addcontentsline{toc}{section}{References}
\end{center}
]

{\footnotesize
\bibliographystyle{ieee_fullname}
\bibliography{egbib}
}

\end{document}